\documentclass{article}
\usepackage{nips15submit_e,times}
\usepackage{hyperref}
\usepackage{url}

\title{Rethinking LDA: Moment Matching for Discrete ICA}

\author{
Anastasia Podosinnikova\qquad Francis Bach\qquad Simon Lacoste-Julien   \\
INRIA - \'Ecole normale sup\'erieure Paris\\
}

\nipsfinalcopy 

\usepackage[parfill]{parskip} 

\usepackage{graphicx}
\usepackage{epstopdf}
\usepackage[font={small}]{caption} 
\usepackage{caption}
\usepackage{subcaption}

\usepackage{color}
\definecolor{mygreen}{RGB}{50, 130, 50}
\hypersetup{
    colorlinks=true, 
    linkcolor=mygreen,  
    citecolor=mygreen,  
    filecolor=mygreen,  
    urlcolor=mygreen,   
}

\makeatletter

\newenvironment{ma}
  {\start@align\@ne\st@rredtrue\m@ne}
  {\endalign}
\makeatother

\newenvironment{mi}{%
  \begin{list}{-}{}
  
}{%
  \end{list}
}

\makeatother

\newcommand\nameeq[2]{\phantom{\text{#2}}&&\begin{gathered}#1\end{gathered}&&\text{#2}}

\newcommand{\emp}[1]{\textbf{#1}}

\usepackage{amssymb,amsmath,amsthm}
\usepackage[makeroom]{cancel} 
\usepackage{dsfont} 

\newtheorem{proposition}{Proposition}[section]

\def\max{\mathop{\rm max}\limits}

\newcommand{\norm}[1]{\left\|#1\right\|}
\newcommand{\normp}[1]{\|#1\|}

\newcommand{\inner}[1]{\left\langle#1\right\rangle}
\newcommand{\innerp}[1]{\langle#1\rangle}

\newcommand{\rbra}[1]{\left(#1\right)}
\newcommand{\sbra}[1]{\left[#1\right]}
\newcommand{\cbra}[1]{\left\{#1\right\}}
\newcommand{\cbras}[1]{\{#1\}}

\newcommand{\wt}[1]{\widetilde{#1}}
\newcommand{\wh}[1]{\widehat{#1}}

\newcommand{\indicator}[1]{\mathds{1}_{\{#1\}}} 
\newcommand{\ones}{\vec{1}}

\newcommand{\diag}{\mathrm{diag}}

\newcommand{\rr}[1]{\mathbb{R}^{#1}}

\newcommand{\pinv}{^{\dagger}}                
\newcommand{\tp}{\otimes}                     

\newcommand{\sumk}{\sum_{k=1}^K}
\newcommand{\summ}{\sum_{m=1}^M}
\newcommand{\sumn}{\sum_{n=1}^N}

\newcommand{\ga}{\alpha}

\newcommand{\gl}{\lambda}
\newcommand{\gt}{\theta}
\newcommand{\gd}{\delta}

\newcommand{\tcal}{\mathcal{T}}
\newcommand{\acal}{\mathcal{A}}
\newcommand{\bcal}{\mathcal{B}}
\newcommand{\ccal}{\mathcal{C}}
\newcommand{\ecal}{\mathcal{E}}
\newcommand{\fcal}{\mathcal{F}}

\newcommand{\ebb}{\mathbb{E}}      
\newcommand{\webb}{\wh{\ebb}}       
\newcommand{\var}{\mathrm{var}}     
      
\newcommand{\cov}{\mathrm{cov}}     
\newcommand{\wcov}{\wh{\cov}}       
\newcommand{\cum}{\mathrm{cum}}    
\newcommand{\wcum}{\wh{\cum}}

\newcommand{\poi}{\mathrm{Poisson}}
\newcommand{\dir}{\mathrm{Dirichlet}}
\newcommand{\gam}{\mathrm{Gamma}}
\newcommand{\mul}{\mathrm{Multinomial}}
\newcommand{\nor}{\mathrm{Normal}}

\usepackage{algorithm}
\usepackage{algcompatible}

\usepackage[numbers]{natbib}
\usepackage{multibib}
\newcites{sup}{Supplementary References} 

\usepackage{tikz}
\usetikzlibrary{arrows,decorations.pathmorphing,fit,positioning}

\makeatletter
\DeclareRobustCommand*\textsubscript[1]{%
  \@textsubscript{\selectfont#1}}
\def\@textsubscript#1{%
  {\m@th\ensuremath{_{\mbox{\fontsize\sf@size\z@#1}}}}}
\makeatother

\newcommand{\rpm}{\raisebox{.2ex}{$\scriptstyle\pm$}}

\begin{document}

\maketitle

\begin{abstract}
\noindent
We consider moment matching techniques for estimation in latent Dirichlet allocation (LDA). By drawing explicit links between LDA and discrete versions of independent component analysis (ICA), we first derive a new set of cumulant-based tensors, with an improved sample complexity. Moreover, we reuse standard ICA techniques such as joint diagonalization of tensors to improve over existing methods based on the tensor power method. In an extensive set of experiments on both synthetic and real datasets, we show that our new combination of tensors and orthogonal joint diagonalization techniques outperforms existing moment matching methods.
\end{abstract}

\section{Introduction}
Topic models have emerged as flexible and important tools for the modelisation of text corpora. While early work has focused on graphical-model approximate inference techniques such as variational inference~\cite{BleEtAl2003} or Gibbs sampling~\cite{Gri2002}, tensor-based moment matching techniques have recently emerged as strong competitors due to their computational speed and theoretical guarantees~\cite{AnaEtAl2012,AnaEtAl2014}. In this paper, we draw explicit links with the independent component analysis (ICA) literature (e.g.,~\cite{ComJut2010} and references therein) by showing a strong relationship between latent Dirichlet allocation (LDA)~\cite{BleEtAl2003} and ICA~\cite{Jut1987,JutHer1991,Com1994}. We can then reuse standard ICA techniques and results, and derive new tensors with better sample complexity and new algorithms based on joint diagonalization.

\section{Is LDA discrete PCA or discrete ICA?}\label{sec:2}
\emp{Notation.} Following the text modeling terminology, we define a corpus $X=\cbra{x_1,\dots,x_N}$ as a collection of $N$ documents. Each document is a collection $\cbra{w_{n1},\dots,w_{nL_n}}$ of $L_n$ tokens. It is convenient to represent the $\ell$-th token of the $n$-th document  as a $1$-of-$M$ encoding with an indicator vector $w_{n\ell}\in\{0,1\}^{M}$ with only one non-zero, where $M$ is the vocabulary size, and each document as the count vector $x_n := \sum_{\ell} w_{n\ell}\in\rr{M}$. In such representation, the length $L_n$ of the $n$-th document is $L_n = \sum_m x_{nm}$. We will always use the index $k\in\cbra{1,\dots,K}$ to refer to topics,the  index $n\in\cbra{1,\dots,N}$ to refer to documents, the index $m\in\cbra{1,\dots,M}$ to refer to words from the vocabulary, and the  index $\ell\in\{1,\dots,L_n\}$ to refer to tokens of the $n$-th document. The plate diagrams of the models from this section are presented in Appendix~\ref{sec:pds}.

\emp{Latent Dirichlet allocation} \cite{BleEtAl2003} is a generative probabilistic model for discrete data such as text corpora.  
In accordance to this model, the $n$-th document is modeled as an \emph{admixture} over the vocabulary of $M$ words with $K$ latent topics. Specifically, the latent variable $\theta_n$, which is sampled from the Dirichlet distribution, represents the topic mixture proportion over $K$ topics for the $n$-th document. Given $\theta_n$,
the topic choice $z_{n\ell} | \theta_n$ for the $\ell$-th token is sampled from the multinomial distribution with the probability vector $\gt_n$. The token $w_{n\ell} | z_{n\ell}, \theta_n$ is then sampled from the multinomial distribution with the probability vector $d_{z_{n\ell}}$, or $d_k$ if $k$ is the index of the non-zero element in $z_{n\ell}$. This vector $d_k$ is the $k$-th topic, that is a vector of probabilities over the words from the vocabulary subject to the simplex constraint, i.e., $d_k\in\Delta_M$, where $\Delta_M := \cbras{d\in\rr{M}\;:\;d\succeq 0,\;\sum_m d_m = 1}$.
This generative process of a document (the index $n$ is omitted for simplicity) can be summarized as
\begin{equation}\label{lda-tokens}
\begin{aligned}
\theta &\sim \dir(c), \\
z_{\ell}|\theta &\sim \mul(1,\theta), \\
w_{\ell} | z_{\ell},\theta &\sim \mul(1,d_{z_{\ell}}).
\end{aligned}
\end{equation}
One can think of the latent variables $z_{\ell}$ as auxiliary variables which were introduced for  convenience of inference, but can in fact be marginalized out \cite{Bun2002}, which leads to the following model
\begin{flalign} \label{lda}
\nameeq{
	\begin{aligned}
	\theta &\sim \dir(c), \\
	x|\theta &\sim \mul(L,D\theta),
	\end{aligned}
}{LDA model}
\end{flalign}
where $D\in\rr{M\times K}$ is the topic matrix with the $k$-th column equal to the $k$-th topic $d_k$, and $c \in \rr{K}_{++}$ is the vector of parameters for the Dirichlet distribution. 
While a document is represented as a set of tokens $w_{\ell}$ in the formulation~\eqref{lda-tokens}, the formulation~\eqref{lda} instead compactly represents a document as the count vector $x$.
Although the two representations are equivalent, we focus on the second one in this paper and therefore refer to it as the LDA model.

Importantly, the LDA model does not model the length of documents. Indeed, although the original paper~\cite{BleEtAl2003} proposes to model the document length as $L|\gl\sim\poi(\gl)$, this is never used in practice and, in particular, the parameter $\gl$ is not learned. 
Therefore, in the way that the LDA model is typically used, it 
does not provide a complete generative process of a document as there is no rule to sample $L|\gl$. 
In this paper, this fact is important, as we need to model the document length in order to make the link with discrete ICA.

\emp{Discrete PCA.} The LDA model~\eqref{lda} can be seen as a discretization of principal component analysis (PCA) via replacement of the normal likelihood with the multinomial one and adjusting the prior~\cite{Bun2002} in the following  probabilistic PCA model~\cite{TipBis1999,Row1998}: $\theta \sim \nor(0,I_K)$ and $x|\theta \sim \nor(D\theta, \sigma^2 I_M)$,
where $D\in\rr{M\times K}$ is a transformation matrix and $\sigma$ is a parameter.

\emp{Discrete ICA (DICA).} Interestingly, a small extension of the LDA model allows its interpretation as a discrete independent component analysis model. 
The extension naturally arises when the document length for the LDA model is modeled as a random variable from the gamma-Poisson mixture (which is equivalent to a negative binomial random variable), i.e., $L|\gl \sim \poi(\gl)$ and $\gl \sim \gam(c_0,b)$, where $c_0 := \sum_k c_k$ is the shape parameter and $b>0$ is the rate parameter. The LDA model~\eqref{lda} with such document length
is equivalent (see Appendix~\ref{sec:ldaproof2}) to 
\begin{flalign}\label{gp}
\nameeq{
	\begin{aligned}
	\ga_k &\sim \gam(c_k,b), \\
	x_m|\ga &\sim \poi([D\ga]_m),
	\end{aligned}
}{GP model}
\end{flalign}
where all $\ga_1,\ga_2,\dots,\ga_K$ are mutually independent, the parameters $c_k$ coincide with the ones of the LDA model in~\eqref{lda}, and the free parameter $b$ can be seen (see Appendix~\ref{sec:L:gp}) as a scaling parameter for the document length when $c_0$ is already prescribed.

This model was introduced by Canny~\cite{Can2004} and later named as a discrete ICA model~\cite{BunJak2004}. 
It is more natural, however, to name model~\eqref{gp} as the gamma-Poisson (GP) model and the model
\begin{flalign}\label{dica}
\nameeq{
	\begin{aligned}
	\ga_1,\dots,\ga_K &\sim \text{mutually independent}, \\
	x_m|\ga & \sim \poi([D\ga]_m)
	\end{aligned}
}{DICA model}
\end{flalign}
 as the discrete ICA (DICA) model. The only difference between~\eqref{dica} and the standard ICA model~\cite{Jut1987,JutHer1991,Com1994} (without additive noise) is the presence of the Poisson noise
which enforces discrete, instead of continuous, values of $x_m$. 
Note also that (a) the discrete ICA model is a \emph{semi-parametric} model that can adapt to any distribution on the topic intensities $\alpha_k$ and that (b) the GP model~\eqref{gp} is a particular case of both the LDA model~\eqref{lda} and the DICA model~\eqref{dica}.

Thanks to this close connection between LDA and ICA, we can reuse standard ICA techniques to derive new efficient algorithms for topic modeling.

\section{Moment matching for topic modeling}
The method of moments estimates latent parameters of a probabilistic model by matching theoretical expressions of its moments with their sample estimates. Recently~\cite{AnaEtAl2012,AnaEtAl2014}, the method of moments was applied to different latent variable models including LDA, resulting in computationally fast learning algorithms with theoretical guarantees. For LDA, they (a) construct {\it LDA moments} with a particular diagonal structure and (b) develop algorithms for estimating the parameters of the model by exploiting this diagonal structure. In this paper, we introduce the novel {\it GP/DICA cumulants} with a similar to the LDA moments structure.
This structure allows to reapply the algorithms of~\cite{AnaEtAl2012,AnaEtAl2014} for the estimation of the model parameters, with the same theoretical guarantees. We also consider another algorithm applicable to both the LDA moments and the GP/DICA cumulants.

\subsection{Cumulants of the GP and DICA models} \label{sec:cum}

In this section, we derive and analyze the novel cumulants of the DICA model. As the GP model is a particular case of the DICA model, all results of this section extend to the GP model.

The first three \emph{cumulant tensors} for the random vector $x$ can be defined as follows
\begin{align}
\label{exp}
\cum(x) & := \ebb(x), \\
\label{covx}
\cum(x,x) & := \cov(x,x) = \ebb\sbra{ (x - \ebb(x)) ( x - \ebb(x))^{\top} }, \\
\label{cumx}
\cum(x,x,x) & := \ebb\sbra{ (x - \ebb(x)) \tp (x - \ebb(x)) \tp (x - \ebb(x))},
\end{align}
where $\tp$ denotes the tensor product (see some properties of  cumulants in Appendix~\ref{sec:cumulants}). The essential property of the cumulants (which does not hold for the moments) that we use in this paper is that the cumulant tensor for a random vector with \emph{independent} components is \emph{diagonal}.

Let $y = D\ga$; then for the Poisson random variable $x_m|y_m \sim \poi (y_m)$, the expectation is $\ebb(x_m|y_m) = y_m$. Hence, by the law of total expectation and the linearity of expectation,  the expectation in~\eqref{exp} has the following form
\begin{equation}
\ebb(x) = \ebb(\ebb(x|y)) = \ebb(y) = D\ebb(\ga).
\end{equation}
Further, the variance of the Poisson random variable $x_m$ is $\var(x_m|y_m) = y_m$ and, as $x_1$, $x_2$,~$\dots$,~$x_M$ are conditionally independent given $y$, then their covariance matrix is diagonal, i.e., $\cov(x,x|y) = \diag(y)$. Therefore, by the law of total covariance, the covariance in~\eqref{covx} has the form
\begin{equation}\label{cum2}
\begin{aligned}
\cov(x,x) & = \ebb\sbra{\cov(x,x|y)} + \cov\sbra{\ebb(x|y),\ebb(x|y)} \\
       & = \diag\sbra{\ebb(y)} + \cov(y,y) 
        = \diag\sbra{\ebb(x)} + D\cov(\ga,\ga)D^{\top},
\end{aligned}
\end{equation}
where the last equality follows by the multilinearity property of cumulants (see Appendix~\ref{sec:cumulants}). Moving the first term from the RHS of~\eqref{cum2} to the LHS, we define
\begin{flalign}\label{S}
\nameeq{
S := \cov(x,x)-\diag\sbra{\ebb(x)}.
}{DICA S-cum.}
\end{flalign}
From~\eqref{cum2} and by the independence of $\ga_1$, $\dots$, $\ga_K$ (see Appendix~\ref{sec:app:dicacum2}),  $S$ has the following diagonal structure
\begin{equation}
\label{diagS}
S = \sum\mathop{}_{k} \var(\ga_k) d_k d_k^{\top} = D\diag\sbra{\var(\ga)}D^{\top}.
\end{equation}

By analogy with the second order case, 
using the law of total cumulance, the multilinearity property of cumulants, and the independence of $\ga_1$, $\dots$, $\ga_K$, we derive in Appendix~\ref{sec:app:3dicacum} the expression~\eqref{cum3}, similar to~\eqref{cum2}, for the third cumulant~\eqref{cumx}. Moving the terms in this expression, we  define a tensor $T$ with the following element
\begin{align}
\phantom{\text{DICA}}& \!\!\!\! \!\!\! \sbra{T}_{m_1m_2m_3} := \cum(x_{m_1},x_{m_2},x_{m_3}) + 2\gd(m_1,m_2,m_3) \ebb(x_{m_1}) \qquad\qquad\, \text{DICA T-cum.} \label{T} \\
& - \gd(m_2,m_3) \cov(x_{m_1},x_{m_2}) - \gd(m_1,m_3) \cov(x_{m_1},x_{m_2}) - \gd(m_1,m_2) \cov(x_{m_1},x_{m_3}), \notag
\end{align}
where $\delta$ is the Kronecker delta.
By analogy with~\eqref{diagS} (Appendix~\ref{sec:app:dicacum2}), the diagonal structure of the tensor $T$:
\begin{equation} 
\label{diagT}
T = \sum\mathop{}_{k} \cum(\ga_k,\ga_k,\ga_k)d_k\tp d_k \tp d_k.
\end{equation}

In Appendix~\ref{sec:lda:moms:notation}, we recall (in our notation) the matrix $S$~\eqref{S:lda} and the tensor $T$~\eqref{T:lda} for the LDA model~\cite{AnaEtAl2012}, which are analogues of the matrix $S$~\eqref{S} and the tensor $T$~\eqref{T} for the GP/DICA models. Slightly abusing terminology, we refer to the matrix $S$~\eqref{S:lda} and the tensor $T$~\eqref{T:lda} as the {\it LDA moments} and to the matrix $S$~\eqref{S} and the tensor $T$~\eqref{T} as the {\it GP/DICA cumulants}. The diagonal structure~\eqref{diagS:lda}~\&~\eqref{diagT:lda} of the LDA moments is similar to the diagonal structure~\eqref{diagS}~\&~\eqref{diagT} of the GP/DICA cumulants, though arising through a slightly different argument, as discussed at the end of Appendix~\ref{sec:lda:moms:notation}. Importantly, due to this similarity, the algorithmic frameworks for both the GP/DICA cumulants and the LDA moments coincide.

The following sample complexity results apply to the sample estimates of the GP cumulants:\footnote{Note that the expected squared error for the DICA cumulants is similar, but the expressions are less compact and, in general, depend on the prior on $\alpha_k$.}
\begin{proposition}\label{sample-complexity}
Under the GP model, the expected error for the sample estimator $\wh{S}$~\eqref{appSgp:fs} for the GP cumulant $S$~\eqref{S} is:
\begin{equation} \label{eq:S_complexity}
\ebb\sbra{ \normp{ \wh{S} - S }_F }  \leq 
\sqrt{\ebb\sbra{ \normp{ \wh{S} - S }_F^2 }}  \le O\rbra{ \frac{1}{\sqrt{N}}  \max\sbra{ \Delta \bar{L}^2,\,\bar{c}_0 \bar{L}  } } ,
\end{equation}
where $\Delta  := \max\mathop{}_k \normp{d_k}_2^2$, $\bar{c}_0 := \min(1,c_0)$ and $\bar{L} := \ebb(L)$.
\end{proposition}
A high probability bound could be derived using  concentration inequalities for  Poisson random variables~\cite{BouEtAl2013}; but the expectation already gives the right order of magnitude for the error (for example via Markov's inequality). 
The expression~\eqref{appSgp:fs} for an unbiased finite sample estimate $\wh{S}$ of $S$ and the expression~\eqref{appTgp:fs} for an unbiased finite sample estimate $\wh{T}$ of $T$ are defined\footnote{For completeness, we also present the finite sample estimates $\wh{S}$~\eqref{S:lda:fs} and $\wh{T}$~\eqref{T:lda:fs} of $S$~\eqref{S:lda} and $T$~\eqref{T:lda} for the LDA moments (which are consistent with the ones suggested in~\cite{AnaEtAl2014}) in Appendix~\ref{sec:lda:empirical}.
} 
in Appendix~\ref{sec:dica:fs}.
A sketch of a proof for Proposition~\ref{sample-complexity} can be found in Appendix~\ref{sec:sample-complexity}.

By following a similar analysis as in~\cite{AnaEtAl2013}, we can rephrase the topic recovery error in term of the error on the GP cumulant. Importantly, the whitening transformation (introduced in Section~\ref{sec:diag}) redivides the error on $S$~\eqref{eq:S_complexity} by $\bar{L}^2$, which is the scale of $S$ (see Appendix~\ref{section-analysis-of-whitening} for details). This means that the contribution from $\hat{S}$ to the recovery error will scale as $O({1}/{\sqrt{N}} \max\{\Delta, {\bar{c}_0}/{\bar{L}}\})$, where both $\Delta$ and ${\bar{c}_0}/{\bar{L}}$ are smaller than $1$ and can be very small. We do not present the exact expression for the expected squared error for the estimator of $T$, but due to a similar structure in the derivation, we expect the analogous bound of $\ebb[ \normp{ \wh{T} - T }_F ] \leq  {1}/{\sqrt{N}}\max\{ \Delta^{3/2} \bar{L}^3,\,\bar{c}_0^{3/2} \bar{L}^{3/2}  \}$.

Current sample complexity results of the LDA moments~\cite{AnaEtAl2012} can be summarized as $O(1/\sqrt{N})$. However, the proof (which can be found in the supplementary material~\cite{AnaEtAl2013}) analyzes only the case when finite sample estimates of the LDA moments are constructed from \emph{one} triple per document, i.e., $w_{1}\tp w_{2} \tp w_{3}$ only, and not from the U-statistics that average multiple (dependent) triples per document as in the practical expressions~\eqref{S:lda:fs} and~\eqref{T:lda:fs} (Appendix~\ref{sec:lda:empirical}). Moreover, one has to be careful when comparing upper bounds. Nevertheless, comparing the bound~\eqref{eq:S_complexity} with the current theoretical results for the LDA moments, we see that the GP/DICA cumulants sample complexity contains the $\ell_2$-norm of the columns of the topic matrix $D$ in the numerator, as opposed to the $O(1)$ coefficient for the LDA moments. This norm can be significantly smaller than $1$ for vectors in the simplex (e.g., $\Delta = O(1/\normp{d_k}_0)$ for sparse topics). This suggests that the GP/DICA cumulants may have better finite sample convergence properties than the LDA moments and our experimental results in Section~\ref{sec:exp:mom} are indeed consistent with this statement.

The GP/DICA cumulants have a somewhat more intuitive derivation than the LDA moments as they are expressed via the count vectors $x$ (which are the sufficient statistics for the model) and not the tokens $w_\ell$'s.
Note also that the construction of the LDA moments depend on the unknown parameter~$c_0$. Given that we are in an unsupervised setting and that moreover the evaluation of LDA is a difficult task~\cite{WalEtAl2009}, setting this parameter is non-trivial. In Appendix~\ref{sec:app:c0lda}, we observe experimentally that the LDA moments are somewhat sensitive to the choice of $c_0$.

\section{Diagonalization algorithms} \label{sec:diag} 

How is the diagonal structure~\eqref{diagS} of $S$ and~\eqref{diagT} of $T$ going to be helpful for the estimation of the model parameters? This question has already been thoroughly investigated in the signal processing (see, e.g.,~\cite{Car1989,Car1990,CarCom1996,Hyv1999,CarSou1993,ComJut2010} and references therein) and machine learning  (see~\cite{AnaEtAl2012,AnaEtAl2014} and references therein) literature. We review the approach in this section. Due to similar diagonal structure, the algorithms of this section apply to both the LDA moments and the GP/DICA cumulants.

For simplicity, let us rewrite the expressions~\eqref{diagS} and~\eqref{diagT} for $S$ and $T$ as follows
\begin{equation}  \label{ST}
\begin{aligned}
S  = \sum\mathop{}_{k} s_k d_k d_k^{\top}, \qquad
T  = \sum\mathop{}_{k} t_k d_k \tp d_k \tp d_k,
\end{aligned}
\end{equation}
where $s_k := \var(\ga_k)$ and $t_k := \cum(\ga_k,\ga_k,\ga_k)$.
Introducing the rescaled topics $\wt{d}_k :=  \sqrt{s_k} d_k$, we can also rewrite
$S = \wt{D} \wt{D}^{\top}$. Following the same assumption from~\cite{AnaEtAl2012} that the topic vectors are linearly independent ($\wt{D}$ is full rank), we can compute a whitening matrix $W\in\rr{K\times M}$ of $S$, i.e., a matrix such that $W S W^{\top} = I_{K}$ where $I_{K}$ is the $K$-by-$K$ identity matrix (see Appendix ~\ref{sec:whitening} for more details). As a result, the vectors $z_k := W\wt{d}_k$ form an orthonormal set of vectors.

Further, let us define a projection
 $\tcal(v)\in\rr{K\times K}$ of a tensor $\tcal\in\rr{K\times K\times K}$  onto a vector $u\in\rr{K}$:
\begin{equation}\label{projT}
\tcal(u)_{k_1 k_2} := \sum\mathop{}_{k_3} \tcal_{k_1k_2k_3} u_{k_3}.
\end{equation}
Applying the multilinear transformation (see, e.g.,~\cite{AnaEtAl2014} for the definition) with $W^{\top}$ to the tensor~$T$ from~\eqref{ST} and projecting the resulting tensor $\tcal :=T(W^{\top},W^{\top},W^{\top})$ onto some vector $u\in\rr{K}$, we obtain
\begin{equation}\label{orthT}
\tcal(u) = \sum\mathop{}_{k} \wt{t}_k \innerp{z_k,u} z_k z_k^{\top},
\end{equation}
where $\wt{t}_k := t_k / s_k^{3/2}$ is due to the rescaling of topics and $\innerp{\cdot,\cdot}$ stands for the inner product.  As the vectors $z_k$ are orthonormal, the pairs $z_k$ and $\gl_k :=\wt{t}_k\innerp{z_k,u}$ are the eigenpairs of the matrix $\tcal(u)$, which are uniquely defined if the eigenvalues $\gl_k$ are all different. If they are unique, we can recover the GP/DICA (as well as LDA) model parameters via $\wt{d}_k = W\pinv z_k$ and $\wt{t}_k = \gl_k / \innerp{z_k,u}$.

This procedure was referred to as
 the spectral algorithm for LDA~\cite{AnaEtAl2012} and the fourth-order\footnote{See Appendix~\ref{sec:on-the-orders-of-cumulants} for a discussion on the orders.}
blind identification algorithm for ICA~\cite{Car1989,Car1990}. Indeed, one can expect that the finite sample estimates
$\wh{S}$~\eqref{appSgp:fs} and $\wh{T}$~\eqref{appTgp:fs}
 possess approximately the diagonal structure~\eqref{diagS} and~\eqref{diagT} and, therefore, the reasoning from above can be applied, assuming that the effect of the sampling error is controlled.

This spectral algorithm, however, is known to be quite unstable in practice (see, e.g.,~\cite{Car1999}).
To overcome this problem, other algorithms were proposed. For ICA, the most notable ones are probably the FastICA algorithm~\cite{Hyv1999} and the JADE algorithm~\cite{CarSou1993}. The FastICA algorithm, with appropriate choice of a contrast function, estimates iteratively the topics, making use of the orthonormal structure~\eqref{orthT}, and performs the deflation procedure at every step. The recently introduced tensor power method (TPM) for the LDA model~\cite{AnaEtAl2014} is close  to the FastICA algorithm. Alternatively, the JADE algorithm modifies the spectral algorithm by performing \emph{multiple} projections  for~\eqref{orthT} and then jointly diagonalizing the resulting matrices with an orthogonal matrix. The spectral algorithm is a special case of this orthogonal joint diagonalization algorithm when only one projection is chosen. Importantly, a fast implementation~\cite{CarSou1996} of the orthogonal joint diagonalization algorithm from~\cite{BunEtAl1993} was proposed, which is based on closed-form iterative Jacobi updates (see, e.g.,~\cite{NocWri2006} for the later).

In practice, the orthogonal joint diagonalization (JD) algorithm is more robust than FastICA (see, e.g.,~\cite[p.~30]{BacJor2002}) or the spectral algorithm. Moreover, although the application of the JD algorithm for the learning of topic models was mentioned in the literature~\cite{AnaEtAl2014,KulEtAl2015}, it was never implemented in practice. In this paper, we apply 
the JD algorithm for the diagonalization of the GP/DICA cumulants as well as the LDA moments, which is described in Algorithm~\ref{alg:jd}. Note that the choice of a projection vector $v_p\in\rr{M}$ obtained as $v_p = \wh{W}^{\top} u_p$ for some vector $u_p \in \rr{K}$ is important and corresponds to the multilinear transformation of $\wh{T}$ with $\wh{W}^{\top}$ along the third mode. Importantly, in Algorithm~\ref{alg:jd}, the joint diagonalization routine is performed over $(P+1)$ matrices of size $K\!\times\!K$, where the number of topics $K$ is usually not too big. This makes the algorithm computationally fast (see Appendix~\ref{sec:code-and-compleixty}). The same is true for the spectral algorithm, but not for TPM.

\begin{algorithm}[th]
   \caption{Joint diagonalization (JD) algorithm for GP/DICA cumulants (or LDA moments)}
   \label{alg:jd}
\begin{algorithmic}[1]
         \STATE \emph{Input:} $X\in{\rr{M\times N}}$, $K$, $P$ (number of random projections); (and $c_0$ for LDA moments)
         \STATE Compute sample estimate $\wh{S}\in\rr{M\times M}$  (\eqref{appSgp:fs} for GP/DICA /~\eqref{S:lda:fs} for LDA in Appendix~\ref{sec:app:implementation})
	\STATE Estimate whitening matrix $\wh{W}\in\rr{K\times M}$ of $\wh{S}$  (see Appendix~\ref{sec:whitening})
	\STATEx 
	\emph{option (a):} Choose vectors $\cbra{u_1,u_2,\dots,u_P}\subseteq\rr{K}$ uniformly at random from the unit $\ell_2$-sphere and set $v_p = \wh{W}^{\top} u_p\in\rr{M}$ for all $p=1,\dots,P$ \quad \quad {\small($P=1$ yields the spectral algorithm)}
	\STATEx \emph{option (b):} Choose vectors $\cbra{u_1,u_2,\dots,u_P}\subseteq\rr{K}$ as the canonical basis $e_1,e_2,\dots,e_K$ of $\rr{K}$ and set $v_p = \wh{W}^{\top} u_p\in\rr{M}$ for all $p=1,\dots,K$
	\STATE For $\forall p$, compute $B_p = \wh{W}\wh{T}(v_p)\wh{W}^{\top}\in\rr{K\times K}$ (\eqref{est:T:dica} for GP/DICA /~\eqref{est:T:lda} for LDA; Appendix~\ref{sec:app:implementation})
	\STATE Perform orthogonal joint diagonalization  of  matrices
	$\cbras{\wh{W}\wh{S}\wh{W}^{\top}=I_K,\; B_p, \;p=1,\dots,P}$ (see~\cite{BunEtAl1993} and~\cite{CarSou1996})
	to find an orthogonal matrix $V\in\rr{K\times K}$ and vectors $\cbra{a_1,a_2,\dots,a_P}\subset\rr{K}$ such that
	$$
	V\wh{W}\wh{S}\wh{W}^{\top} V^{\top} = I_K, \;\mbox{and}\; VB_pV^{\top} \approx\diag(a_p), \; p =1,\dots,P
	$$
	\STATE Estimate joint diagonalization matrix $A=V \wh{W}$ and values $a_p$, $p=1,\dots,P$
	\STATE \emph{Output:} Estimate of $D$ and $c$ as described in Appendix~\ref{estimation-model-parameters} 
\end{algorithmic}
\end{algorithm}

In Section~\ref{sec:diagcmp}, we compare experimentally the performance of the spectral, JD, and TPM algorithms for the estimation of the parameters of the GP/DICA as well as LDA models. We are not aware of any experimental comparison of these algorithms in the LDA context. 
While already working on this manuscript, the JD algorithm was also independently analyzed by~\cite{KulEtAl2015} in the context of tensor factorization for general latent variable models.
However,~\cite{KulEtAl2015} focused mostly on the comparison of approaches for tensor factorization and their stability properties, with brief experiments using a latent variable model related but not equivalent to LDA for community detection. 
In contrast, we provide a detailed experimental comparison in the context of LDA in this paper, as well as propose a novel cumulant-based estimator.
Due to the space restriction 
the estimation of the topic matrix $D$ and the (gamma/Dirichlet) parameter $c$ are moved to  Appendix~\ref{estimation-model-parameters}.

\section{Experiments}\label{sec:exps}
In this section, (a) we compare experimentally the GP/DICA cumulants with the LDA moments and (b)
the spectral algorithm~\cite{AnaEtAl2012}, the tensor power method~\cite{AnaEtAl2014} (TPM), the joint diagonalization (JD) algorithm from Algorithm~\ref{alg:jd},
 and variational inference for LDA~\cite{BleEtAl2003}.

\emp{Real data:} the associated press (AP) dataset, from D.~Blei's web page,\footnote{\url{http://www.cs.columbia.edu/~blei/lda-c}} with $N = 2,243$ documents and $M = 10,473$ vocabulary words and the average document length $\wh{L}=194$; the NIPS papers dataset\footnote{\url{http://ai.stanford.edu/~gal/data}}~\cite{nips_data} of $2,483$ NIPS papers and $14,036$ words, and $\wh{L}=1,321$; the KOS dataset,\footnote{\url{https://archive.ics.uci.edu/ml/datasets/Bag+of+Words}} from the UCI Repository, with $3,430$ documents and $6,906$ words, and $\wh{L} = 136$.

\emp{Semi-synthetic data} are constructed 
by analogy with~\cite{AroEtAl2013}: (1) the LDA parameters $D$ and~$c$ are learned from the real datasets with variational inference and (2) toy data are sampled from a model of interest with the given parameters $D$ and~$c$. This provides the ground truth parameters $D$ and~$c$. For each setting, data are sampled 5 times and the results are averaged. We plot error bars that are the minimum and maximum values. For the AP data, $K\in\{10,50\}$ topics are learned and, for the NIPS data, $K\in\{10,90\}$ topics are learned. 
For larger $K$, the obtained topic matrix is ill-conditioned, which violates the identifiability condition for topic recovery using moment matching techniques~\cite{AnaEtAl2012}. All the documents with less than $3$ tokens are resampled.

\emp{Sampling techniques.} All the sampling models have the parameter $c$ which is set to $c = c_0 \bar{c}/\norm{\bar{c}}_1$, where $\bar{c}$ is the learned $c$ from the real dataset with 
variational LDA, and $c_0$ is a parameter that we can vary.
The {\textit{GP}} data are sampled from the gamma-Poisson model~\eqref{gp} with $b=c_0/\wh{L}$ so that the expected document length is $\wh{L}$ (see Appendix~\ref{sec:L:gp}).
The {\textit{LDA-fix($L$)}} data are sampled from the LDA model~\eqref{lda} with the document length being fixed to a given $L$. The {\textit{LDA-fix2($\gamma$,$L_1$,$L_2$)}} data are sampled as follows: 
 $(1-\gamma)$-portion  of the documents are sampled from the \textit{LDA-fix($L_1$)} model with a given document length $L_1$ and $\gamma$-portion of the documents are sampled from the \textit{LDA-fix($L_2$)} model with a given document length $L_2$.

\emp{Evaluation.} The evaluation of topic recovery for semi-synthetic data is performed with the $\ell_1$-error between the recovered $\wh{D}$ and true $D$ topic matrices with the best permutation of columns:
$\text{err}_{\ell_1} (\wh{D},D) := \min_{\pi \in \mathrm{PERM}} \frac{1}{2K} \sum\mathop{}_{k} \| \wh{d}_{\pi_k} - d_{k}\|_1 \;\in [0,1]$.
The minimization is over the possible permutations $\pi \in \mathrm{PERM}$ of the columns of 
$\wh{D}$ and can be efficiently obtained with the Hungarian algorithm for bipartite matching.
For the evaluation of topic recovery in the real data case, we use an approximation of the log-likelihood for held out documents as the metric~\cite{WalEtAl2009}.
See Appendix~\ref{sec:evaluation-real} for more details.

We use our Matlab implementation of the GP/DICA cumulants, the LDA moments, and the diagonalization algorithms. The datasets and the code for reproducing our experiments are available online.\footnote{ \url{https://github.com/anastasia-podosinnikova/dica}} In Appendix~\ref{sec:code-and-compleixty}, we discuss the complexity and implementation of the algorithms.
We explain how we initialize the parameter $c_0$ for the LDA moments in Appendix~\ref{sec:initialization-c0}.

\subsection{Comparison of the diagonalization algorithms}\label{sec:diagcmp}
\begin{figure}[t]
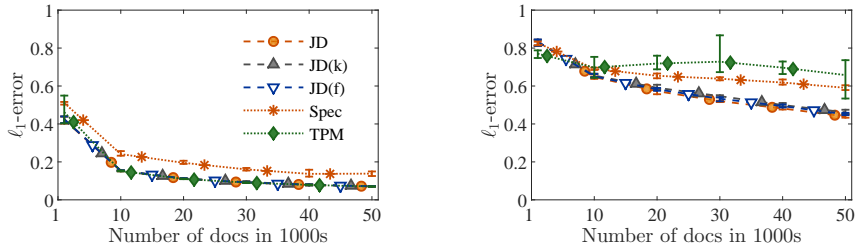

\centering
\begin{tabular}{cccc}
\includegraphics[clip=true, trim=0em 1em 0em 1em,width=.36\columnwidth]{fig_1_left_l1.eps} 
 & 
 
 &
 
 &
\includegraphics[clip=true, trim=0em 1em 0em 1em,width=.36\columnwidth]{fig_1_right_l1.eps} 
\end{tabular}
\vspace{-1em}
\caption{
Comparison of the diagonalization algorithms. 
The topic matrix $D$ and Dirichlet parameter $c$ are learned for $K=50$ from AP; $c$ is scaled to sum up to $0.5$ 
and $b$ is set to fit the expected document length $\wh{L}=200$. The semi-synthetic dataset is sampled from \textit{GP}; number of documents $N$ varies from $1,000$ to $50,000$. \emp{Left:} GP/DICA moments. \emp{Right:} LDA moments.  \textit{Note}: a smaller value of the $\ell_1$-error is better. 
}
\label{plots:diag}
\vspace{-1.5em}
\end{figure}

In Figure~\ref{plots:diag}, we compare the diagonalization algorithms on the semi-synthetic AP dataset for $K=50$ using the GP  sampling. We compare the tensor power method (TPM)~\cite{AnaEtAl2014}, the spectral algorithm (Spec), the orthogonal joint diagonalization algorithm (JD) described in Algorithm~\ref{alg:jd} with different options to choose the random projections:
JD(k) takes $P=K$ vectors $u_p$ sampled uniformly from the unit $\ell_2$-sphere in $\rr{K}$ and selects $v_p = W^{\top} u_p$ (option (a) in Algorithm~\ref{alg:jd}); JD selects the full basis $e_1,\dots,e_K$ in $\rr{K}$ and sets $v_p = W^{\top} e_p$ (as JADE~\cite{CarSou1993}) (option (b) in Algorithm~\ref{alg:jd}); $JD(f)$ chooses the full canonical basis of $\rr{M}$ as the projection vectors (computationally expensive). 

Both the GP/DICA cumulants and LDA moments are well-specified in this setup. However, the LDA moments have a slower finite sample convergence and, hence, a larger estimation error for the same value $N$. As expected, the spectral algorithm is always slightly inferior to the joint diagonalization algorithms. With the GP/DICA cumulants, where the estimation error is low, all algorithms demonstrate good performance, which also fulfills our expectations. 
However, although TPM shows almost perfect performance in the case of the GP/DICA cumulants (left), it significantly deteriorates for the LDA moments (right), which can be explained by the larger estimation error of the LDA moments and lack of robustness of TPM. 
The running times are discussed in Appendix~\ref{sec:runtimes}.
Overall, the orthogonal joint diagonalization algorithm with initialization of random projections as $W^{\top}$ multiplied with the canonical basis in $\rr{K}$ (JD) is both computationally efficient and fast.

\subsection{Comparison of the GP/DICA cumulants and the LDA moments}\label{sec:exp:mom}
\begin{figure}[!h]
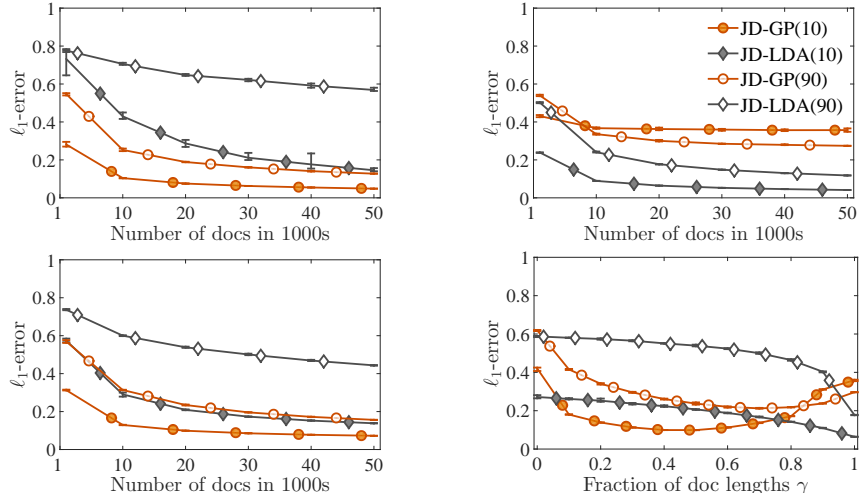

\centering
\begin{tabular}{cccc}
\includegraphics[clip=true, trim=0em 1em 0em 1em,width=.36\columnwidth]{fig_2_top_left_l1.eps} 
 & 
 
 &
 
 &
\includegraphics[clip=true, trim=0em 1em 0em 1em,width=.36\columnwidth]{fig_2_top_right_l1.eps} 
\\ 
\includegraphics[clip=true, trim=0em 1em 0em 1em,width=.36\columnwidth]{fig_2_bottom_left_l1.eps}  
&

&

&
\includegraphics[clip=true, trim=0em 1em 0em 1em,width=.36\columnwidth]{fig_2_bottom_right_l1.eps} 
\end{tabular}
\vspace{-1em}
\caption{ Comparison of the GP/DICA cumulants and LDA moments. Two topic matrices and parameters $c_1$ and $c_2$ are learned from the NIPS dataset for $K=10$ and $90$; $c_1$ and $c_2$ are scaled to sum up to $c_0=1$.  Four corpora of different sizes $N$ from $1,000$ to $50,000$: \emp{top, left:} $b$ is set to fit the expected document length $\wh{L} = 1300$; sampling from the \textit{GP} model; \emp{top, right:} sampling from the \textit{LDA-fix(200)} model; \emp{bottom, left:} sampling from the \textit{LDA-fix2(0.5,20,200)} model. \emp{Bottom, right:} the number of documents here is fixed to $N = 20,000$; sampling from the \textit{LDA-fix2($\gamma$,20,200)} model varying the values of the fraction $\gamma$ from $0$ to $1$ with the step~$0.1$. \textit{Note}: a smaller value of the $\ell_1$-error is better. }
\label{plot:moms}
\vspace{-1em}
\end{figure}

In Figure~\ref{plot:moms}, when sampling from the \textit{GP} model (top, left), both the GP/DICA cumulants and LDA moments are well specified, which implies that the approximation error (i.e., the error w.r.t. the model (mis)fit) is low for both. The GP/DICA cumulants achieve low values of the estimation error already for $N=10,000$ documents independently of the number of topics, while the convergence is slower for the LDA moments.
When sampling from the \textit{LDA-fix(200)} model (top, right), the GP/DICA cumulants are mis-specified and their approximation error is high, although the estimation error is low due to the faster finite sample convergence.  One reason of poor performance of the GP/DICA cumulants, in this case, is the absence of variance in the document length. Indeed, if documents with two different lengths are mixed by sampling from the \textit{LDA-fix2(0.5,20,200)} model (bottom, left), the GP/DICA cumulants performance improves. Moreover, the experiment with a changing fraction $\gamma$ of documents (bottom, right) shows that a non-zero variance on the length improves the performance of the GP/DICA cumulants. As in practice real corpora usually have a non-zero variance for the document length, this bad scenario for the GP/DICA cumulants is not likely to happen.

\subsection{Real data experiments}\label{section-real-experiments}
\begin{figure}[t!]
\begin{center}
\begin{tabular}{cccc}
\includegraphics[clip=true, trim=0em 1em 0em 1em,width=.36\columnwidth]{fig_3_left_ap.eps} 
 & 
\includegraphics[width=.14\columnwidth]{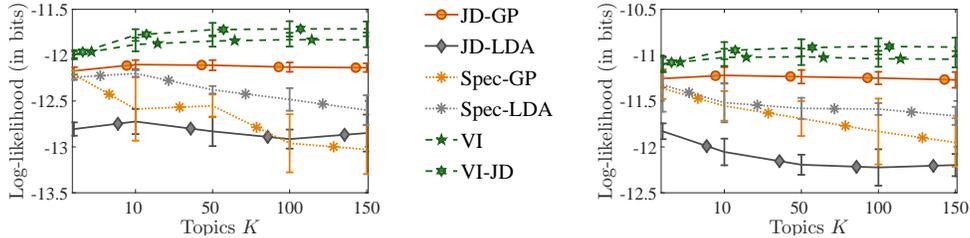} 
 & 
\includegraphics[clip=true, trim=0em 1em 0em 1em,width=.36\columnwidth]{fig_3_right_kos.eps} 
\end{tabular}
\vspace{-1em}
\caption{ 
Experiments with real data. \emp{Left:} the AP dataset. \emp{Right:} the KOS dataset. \textit{Note}: a higher value of the log-likelihood is better.
}
\label{lastfigure}
\end{center}
\vspace{-2em}
\end{figure}

In Figure~\ref{lastfigure}, JD-GP, Spec-GP, JD-LDA, and Spec-LDA are compared with variational inference (VI) and with variational inference initialized with the output of JD-GP (VI-JD). We measure the held out log-likelihood per token (see Appendix~\ref{sec-expsup-real} for details on the experimental setup). The orthogonal joint diagonalization algorithm with the GP/DICA cumulants (JD-GP) demonstrates promising performance.
In particular, the GP/DICA cumulants significantly outperform the LDA moments. Moreover, although variational inference performs better than the JD-GP algorithm, restarting variational inference with the output of the JD-GP algorithm systematically leads to better results. Similar behavior has already been observed (see, e.g.,~\cite{CohCol2014}).

\section{Conclusion}
In this paper, we have proposed a new set of tensors for a discrete ICA model related to LDA, where word counts are directly modelled. These moments make fewer assumptions regarding distributions, and are theoretically and empirically more robust than previously proposed tensors for LDA, both on synthetic and real data. Following the ICA literature, we showed that our joint diagonalization procedure is also more robust. Once the topic matrix has been estimated in a semi-parametric way where topic intensities are left unspecified, it would be interesting to learn the unknown distributions of the independent topic intensities.

\emp{Aknowledgements.} This work was partially supported by the MSR-Inria Joint Center. The authors would like to thank Christophe Dupuy for helpful discussions.

\subsubsection*{References}
\renewcommand\refname{\vspace{-2em}}
\bibliographystyle{unsrt}
{\small\bibliography{lit}}

\newpage
\appendix
\renewcommand{\thesubsection}{\Alph{subsection}}
\subsection{Appendix. Plate diagrams for the models from Section~\ref{sec:2}}\label{sec:pds}

\begin{figure}[htp]
\begin{subfigure}{.24\textwidth}
\centering
\begin{tikzpicture}
[
observed/.style={minimum size=28pt,circle,draw=black,fill=black!10},
unobserved/.style={minimum size=28pt,circle,draw},
arrow/.style={->,>=stealth',semithick},
]
\node (wnl) [observed] at (0,0) {$w_{n\ell}$};
\node (znl) [unobserved] at (0,1.4) {$z_{n\ell}$};
\node (theta) [unobserved] at (0,3.3) {$\theta_n$};
\node (c) [label=above:$c$] at (0,4.9) {};
\filldraw [black] (0,4.9) circle (3pt);
\node (D) [label=above:$D$,inner sep=2pt,outer sep=0pt] at (-1.5,1.5) {};
\filldraw [black] (-1.5,1.5) circle (3pt);
\path
(znl)   edge [arrow] (wnl)
(c)     edge [arrow] (theta)
(theta) edge [arrow] (znl)
(D)     edge [arrow] (wnl)
;
\node [draw,fit=(wnl) (theta), inner sep=18pt] (plate-context) {};
\node [below left] at (plate-context.north east) {$N$};
\node [draw,fit=(wnl) (znl), inner sep=14pt] (plate-token) {};
\node [below left] at (plate-token.north east) {$L_n$};
\end{tikzpicture}
\caption{LDA~\eqref{lda-tokens-n}}\label{pd-lda-tokens}
\end{subfigure}%
\begin{subfigure}{.24\textwidth}
\centering
\begin{tikzpicture}
[
observed/.style={minimum size=28pt,circle,draw=black,fill=black!10},
unobserved/.style={minimum size=28pt,circle,draw},
arrow/.style={->,>=stealth',semithick},
]
\node (xn) [observed] at (0,0) {$x_n$};
\node (theta) [unobserved] at (0,3.3) {$\theta_n$};
\node (c) [label=above:$c$] at (0,4.9) {};
\filldraw [black] (0,4.9) circle (3pt);
\node (D) [label=above:$D$,inner sep=2pt,outer sep=0pt] at (-1.5,1.5) {};
\filldraw [black] (-1.5,1.5) circle (3pt);
\path
(c)     edge [arrow] (theta)
(theta) edge [arrow] (xn)
(D)     edge [arrow] (xn)
;
\node [draw,fit=(xn) (theta), inner sep=18pt] (plate-context) {};
\node [below left] at (plate-context.north east) {$N$};
\end{tikzpicture}
\caption{LDA~\eqref{lda-n}}\label{pd-lda} 
\end{subfigure}%
\begin{subfigure}{.24\textwidth}
\centering
\begin{tikzpicture}
[
observed/.style={minimum width=28pt,circle,draw=black,fill=black!10},
unobserved/.style={minimum size=28pt,circle,draw},
arrow/.style={->,>=stealth',semithick},
]
\node (xn) [observed] at (0,0) {$x_{nm}$};
\node (alpha) [unobserved] at (0,3.3) {$\alpha_n$};
\node (c) [label=above:$c$] at (0,4.9) {};
\filldraw [black] (0,4.9) circle (3pt);
\node (D) [label=above:$D$,inner sep=2pt,outer sep=0pt] at (-1.5,1.5) {};
\filldraw [black] (-1.5,1.5) circle (3pt);
\path
(c)     edge [arrow] (alpha)
(alpha) edge [arrow] (xn)
(D)     edge [arrow] (xn)
;
\node [draw,fit=(xn) (alpha), inner sep=18pt] (plate-context) {};
\node [below left] at (plate-context.north east) {$N$};
\node [draw,fit=(xn), inner sep=14pt] (plate-context) {};
\node [below left] at (plate-context.north east) {$M$};
\end{tikzpicture}
\caption{GP~\eqref{gp-n}}\label{pd-gp} 
\end{subfigure} 
\begin{subfigure}{.24\textwidth}
\centering
\begin{tikzpicture}
[
observed/.style={minimum width=28pt,circle,draw=black,fill=black!10},
unobserved/.style={minimum size=28pt,circle,draw},
arrow/.style={->,>=stealth',semithick},
]
\node (xn) [observed] at (0,0) {$x_{nm}$};
\node (alpha) [unobserved] at (0,3.3) {$\alpha_n$};
\node (D) [label=above:$D$,inner sep=2pt,outer sep=0pt] at (-1.5,1.5) {};
\node (c) [label=above:$ $] at (0,4.9) {};
\filldraw [white] (0,4.9) circle (3pt);
\filldraw [black] (-1.5,1.5) circle (3pt);
\path
(alpha) edge [arrow] (xn)
(D)     edge [arrow] (xn)
;
\node [draw,fit=(xn) (alpha), inner sep=18pt] (plate-context) {};
\node [below left] at (plate-context.north east) {$N$};
\node [draw,fit=(xn), inner sep=14pt] (plate-context) {};
\node [below left] at (plate-context.north east) {$M$};
\end{tikzpicture}
\caption{DICA~\eqref{dica-n}}\label{pd-dica} 
\end{subfigure} 
\caption{Plate diagrams for the models from Section~\ref{sec:2}.} 
\end{figure}
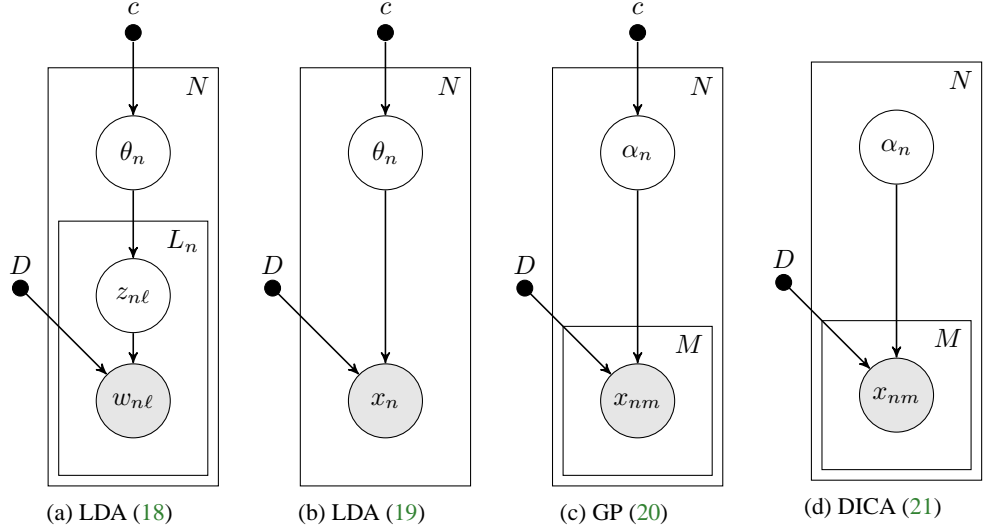

In Section~\ref{sec:2}, the index $n$, which stands for the $n$-th document, was omitted. For convenience, we recall the models.
The LDA model in the tokens representation:
\begin{equation}\label{lda-tokens-n}
  \begin{aligned}
    \theta_n &\sim \dir(c), \\
    z_{n\ell}|\theta_n &\sim \mul(1,\theta_n), \\
    w_{n\ell} | z_{n\ell},\theta_n &\sim \mul(1,d_{z_{n\ell}});
  \end{aligned}
\end{equation}
the LDA model with the marginalized out latent variable $z$:
\begin{equation} \label{lda-n}
  \begin{aligned}
    \theta_n &\sim \dir(c), \\
    x_n|\theta_n &\sim \mul(L_n,D\theta_n);
  \end{aligned}
\end{equation}
the GP model:
\begin{equation}\label{gp-n}
  \begin{aligned}
	\ga_{nk} &\sim \gam(c_k,b), \\
	x_{nm}|\ga_n &\sim \poi([D\ga_n]_m);
  \end{aligned}
\end{equation}
and the DICA model:
\begin{equation}\label{dica-n}
  \begin{aligned}
	\ga_{n1},\dots,\ga_{nK} &\sim \text{mutually independent}, \\
	x_{nm}|\ga_n & \sim \poi([D\ga_n]_m).
  \end{aligned}
\end{equation}

\subsection{Appendix. The GP model}
\subsubsection{The connection between the LDA and GP models} \label{sec:ldaproof2}
To show that the LDA model~\eqref{lda} with the additional assumption that the document length is modeled as a gamma-Poisson random variable is equivalent to the GP model~\eqref{gp}, we show that:
\begin{mi}
\item when modeling the document length $L$ as a Poisson random variable with a parameter $\gl$, the count vectors $x_{1}$, $x_{2}$, $\dots$, $x_{M}$ are mutually independent Poisson random variables;
\item the Gamma prior on $\gl$ reveals the connection $\ga_{k} = \gl \gt_{k}$ between the Dirichlet random variable $\gt$ and the mutually independent gamma random variables $\ga_{1}$, $\ga_{2}$, $\dots$, $\ga_{K}$.
\end{mi}

For completeness, we repeat the known result that if $L \sim \poi(\lambda)$ and $x | L \sim \mul(L,D\theta)$ (which thus means that $L = \sum_m x_{m}$ with probability one), then $x_{1}$, $x_{2}$, $\dots$, $x_{M}$ are mutually independent Poisson random variables with parameters $\gl\sbra{D\gt}_1$, $\gl\sbra{D\gt}_2$, $\dots$, $\gl\sbra{D\gt}_M$. Indeed, we consider the following joint probability mass function where $x$ and $L$ are assumed to be non-negative integers:
\begin{ma}
p(x, L|\gt,\gl) =& p(L|\gl) p(x|L,\gt) \\
=& \indicator{ L=\sum_m x_{m} } \, \frac{\exp\rbra{-\gl}\gl^{L}}{\cancel{L!}}\frac{\cancel{L!}}{\prod_m x_{m}!} \prod_m \sbra{D\gt}_m^{x_{m}}  \\
=& \indicator{ L=\sum_m x_{m}} \, \exp (-\gl \sum_m \sbra{D\gt}_m) \gl^{\sum_m x_{m}} \prod_m \frac{ \sbra{D\gt}_m^{x_{m}}}{x_{m}!}  \\
=& \indicator{ L=\sum_m x_{m}} \, \prod_m \frac{ \exp(-\gl\sbra{D\gt}_m) (\gl \sbra{D\gt}_m)^{x_{m}} } {x_{m}!} \\
=& \indicator{ L=\sum_m x_{m}} \, \prod_m \poi(x_{m}; \gl \sbra{D\gt}_m),
\end{ma}
where in the third equation we used the fact that 
$$\sum_m \sbra{D\gt}_m = \sum_{m,k}D_{mk}\gt_k=\sum_k\gt_k \sum_mD_{mk}=1.$$ 
We thus have $p(x,L | \theta, \lambda) = p(L|x) \prod_{m} p(x_m | \lambda [D\theta]_m)$ where $p(L|x)$ is simply the deterministic distribution $\indicator{ L=\sum_m x_{m}}$ and $p(x_m | \lambda [D\theta]_m)$ for $m=1, \ldots, M$ are independent $\poi(\lambda [D\theta]_m)$ distributions (and thus do not depend on $L$).
Note that in the notation introduced in the paper, $D_{mk} = d_{km}$.
Hence, by using the construction of the Dirichlet distribution from the normalization of independent gamma random variables, we can show that the LDA model with a gamma-Poisson prior over the length is equivalent to the following model (recall, that  $c_0 = \sum_k  c_k$):
\begin{equation}\label{intermodel}
\begin{aligned}
\gl \sim & \;\gam(c_0,b), \\
\gt \sim& \;\dir(c), \\
x_{m}|\gl, \theta \sim &\; \poi(\sbra{D(\gl\gt)}_m).
\end{aligned}
\end{equation}

More specifically, we complete the second part of the argument with the following properties. When $\ga_{1}$, $\ga_{2}$, $\dots$, $\ga_{K}$ are mutually independent gamma random variables, each $\ga_{k}\sim\gam(c_k,b)$, their sum is also a gamma random variable $\sum_k \ga_{k} \sim\gam(\sum_k c_k,b)$. The former is equivalent to $\gl$. 
It is known (e.g.,~\citesup{FriEtAl2010}) that a Dirichlet random variable can be sampled by first sampling independent gamma random variables ($\ga_k$) and then dividing each of them by their sum ($\gl$): $\theta_k = \ga_k / \sum_{k'} \ga_{k'}$, and, in other direction, the variables $\alpha_k = \gl \theta_k$ are mutually independent, giving back the GP model~\eqref{gp}.

\subsubsection{The expectation and the variance of the document length for the GP model}\label{sec:L:gp}
From the drivations in Appendix~\ref{sec:ldaproof2}, it follows that the document length of the GP model~\eqref{gp} is a gamma-Poisson random variable, i.e., $L|\gl \sim\poi(\gl)$ and $\gl \sim \gam(c_0,b)$. Therefore, the following follows from the law of total expectation and the law of total variance
\begin{ma}
\ebb(L) &=\ebb\sbra{\ebb(L|\gl)} = \ebb(\gl)=c_0/b \\
\var(L) &= \var \sbra{\ebb(L|\gl)} + \ebb\sbra{\var(L|\gl)}=\var(\gl)+\ebb(\gl) = c_0/b+c_0/b^2
\end{ma}
The first expression shows that the parameter $b$ controls the expected document length $\ebb(L)$ for a given parameter $c_0$: the smaller $b$, the larger $\ebb(L)$. On the other hand, if we allow $c_0$ to vary as well, only the ratio $c_0/b$ is important for the document length. We can then interpret the role of $c_0$ as actually controlling the concentration of the distribution for the length $L$ (through the variance). More specifically, we have that:
\begin{equation} \label{eq:concentrationL_GP}
\frac{\var(L)}{(\ebb(L))^2} = \frac{1}{\ebb(L)} + \frac{1}{c_0}.
\end{equation}
For a fixed target document length $\ebb(L)$, we can increase the variance (and thus decrease the concentration) by using a smaller $c_0$.

\subsection{Appendix. The cumulants of the GP and DICA models} \label{sec:app:dicacum}
\subsubsection{Cumulants}\label{sec:cumulants}
For a random vector $x\in\rr{M}$,  the first three cumulant tensors\footnote{Strictly speaking, the (scalar) $n$-th cumulant $\kappa_n$ of a random variable $X$ is defined via the cumulant-generating function $g(t)$, which is the natural logarithm of the moment-generating function, i.e $g(t) := \log\ebb\sbra{e^{tX}}$. The cumulant $\kappa_n$ is then obtained from a power series expansion of the cumulant-generating function, that is $g(t) = \sum_{n=1}^{\infty}\kappa_n{t^n}/{n!}$ [Wikipedia].} are
\begin{ma}
\cum(x_m) &= \ebb(x_m), \\
\cum(x_{m_1},x_{m_2}) &= \ebb\sbra{(x_{m_1}-\ebb(x_{m_1}))(x_{m_2}-\ebb(x_{m_2})} = \cov(x_{m_1},x_{m_2}), \\
\cum(x_{m_1},x_{m_2},x_{m_3}) &= \ebb\sbra{(x_{m_1}-\ebb(x_{m_1}))(x_{m_2}-\ebb(x_{m_2}
))(x_{m_3}-\ebb(x_{m_3})))}.
\end{ma}
Note that the 2nd and 3rd cumulants coincide with the 2nd and 3rd central moments (but not for higher orders).
In the following, $\cum(x,x,x)\in\rr{M \times M\times M}$ denotes the third order tensor with elements $\cum(x_{m_1},x_{m_2},x_{m_3})$.
Some of the properties of cumulants are listed below (see~\cite[chap.~5]{ComJut2010}). The most important property that motivate us to use cumulants in this paper (and the ICA literature) is the \textbf{independence} property, which says that the cumulant tensor for a random vector with independent components is diagonal (this property \emph{does not} hold for the (non-central) moment tensors of any order, and neither for the central moments of order 4 or more).
\begin{mi}
\item \emp{Independence.} If the elements of $x\in\rr{M}$ are independent, then their cross-cumulants are zero as soon as two indices are different, i.e., $\cum(x_{m_1},x_{m_2})=\gd(m_1,m_2)\ebb[(x_{m_1}-\ebb_{m_1}))^2]$ and $\cum(x_{m_1},x_{m_2},x_{m_3})=\gd(m_1,m_2,m_3)\ebb[(x_{m_1}-\ebb(x_{m_1}))^3]$, where $\gd$ is the Kronecker delta.
\item \emp{Multilinearity.} If two random vectors $y\in\rr{M}$ and $\ga\in\rr{K}$ are linearly dependent, i.e., $y=D\ga$ for some $D\in\rr{M\times K}$, then
\begin{ma}
\cum(y_m) & = \sum_k \cum(\ga_k) D_{mk},\\
\cum(y_{m_1},y_{m_2}) &= \sum_{k_1,k_2} \cum(\ga_{k_1},\ga_{k_2}) D_{m_1k_1}D_{m_2k_2},\\
\cum(y_{m_1},y_{m_2},y_{m_3}) &= \sum_{k_1,k_2,k_3} \cum(\ga_{k_1},\ga_{k_2},\ga_{k_3}) D_{m_1k_1}D_{m_2k_2}D_{m_3k_3},
\end{ma}
which can also be denoted\footnote{In~\cite{AnaEtAl2014}, given a tensor $T\in\rr{K\times K\times K}$, $T(D^{\top},D^{\top},D^{\top})$ is referred to as the multilinear map. In~\citesup{KolBad2009}, the same entity is denoted by $T\times_1 D^{\top} \times_2 D^{\top} \times_3 D^{\top}$, where $\times_n$ denotes the $n$-mode tensor-matrix product.} 
by 
\begin{ma}
\ebb(y) &= D\ebb(\ga),\\
\cov(y,y) &=D\cov(\ga,\ga)D^{\top},\\
 \cum(y,y,y)&=\cum(\ga,\ga,\ga)(D^{\top},D^{\top},D^{\top}).
\end{ma}
\item \emp{The law of total cumulance.} For two random vectors $x\in\rr{M}$ and $y\in\rr{M}$, it holds
\begin{ma}
\cum(x_m) &= \ebb\sbra{\ebb(x_m|y)}, \\
\cum(x_{m_1},x_{m_2}) &=  \ebb\sbra{\cov(x_{m_1},x_{m_2}|y)} + \cov\sbra{\ebb(x_{m_1}|y),\ebb(x_{m_2}|y)}, \\
\cum(x_{m_1},x_{m_2},x_{m_3}) &= \ebb\sbra{\cum(x_{m_1},x_{m_2},x_{m_3}|y)} + \cum\sbra{\ebb(x_{m_1}|y),\ebb(x_{m_2}|y),\ebb(x_{m_3}|y)} \\
&+ \cov\sbra{\ebb(x_{m_1}|y),\cov(x_{m_2},x_{m_3}|y)} \\
&+ \cov\sbra{\ebb(x_{m_2}|y),\cov(x_{m_1},x_{m_3}|y)} \\
&+ \cov\sbra{\ebb(x_{m_3}|y),\cov(x_{m_1},x_{m_2}|y)}.
\end{ma}
Note that the first expression is also well known as the law of total expectation or the tower property, while the second one is known as the law of total covariance.
\end{mi}

\subsubsection{The third cumulant of the GP/DICA models} \label{sec:app:3dicacum}
In this section, by analogy with Section~\ref{sec:cum}, we derive the third GP/DICA cumulant.

As the third cumulant of a Poisson random variable $x_m$ with parameter $y_m$ is $\ebb((x_{m}-\ebb(x_{m}))^3|y_{m}) = y_{m}$, then by the independence property of cumulants from Section~\ref{sec:cumulants}, the cumulant of $x|y$ is diagonal:
$$
\cum(x_{m_1},x_{m_2},x_{m_3}|y)=\gd(m_1,m_2,m_3)\, y_{m_1}.
$$
Substituting the cumulant of $x|y$ into the law of total cumulance, we obtain
\begin{align}
&\cum( x_{m_1}, x_{m_2},x_{m_3}) = \ebb\sbra{\cum(x_{m_1},x_{m_2},x_{m_3}|y)} \notag \\
 & \quad+ \cum\sbra{\ebb(x_{m_1}|y),\ebb(x_{m_2}|y),\ebb(x_{m_3}|y)} 
  + \cov\sbra{\ebb( x_{m_1}|y),\cov(x_{m_2},x_{m_3}|y)}  \notag \\
 &\quad + \cov\sbra{\ebb(x_{m_2}|y),\cov(x_{m_1},x_{m_3}|y)} 
  + \cov\sbra{\ebb(x_{m_3}|y),\cov(x_{m_1},x_{m_2}|y)}  \notag \\
 &= \gd(m_1,m_2,m_3) \ebb(y_{m_1}) + \cum(y_{m_1},y_{m_2},y_{m_3})  \notag \\
 &\quad+ \gd(m_2,m_3)\cov(y_{m_1},y_{m_2}) 
 + \gd(m_1,m_3) \cov(y_{m_1},y_{m_2}) 
 + \gd(m_1,m_2) \cov(y_{m_1},y_{m_3})  \notag \\
 &= \gd(m_1,m_2,m_3) \ebb(x_{m_1}) + \cum(y_{m_1},y_{m_2},y_{m_3})  \notag \\
 &\quad + \gd(m_2,m_3) \cov(x_{m_1},x_{m_2}) - \gd(m_1,m_2,m_3) \ebb(x_{m_1})  \notag \\
 &\quad + \gd(m_1,m_3) \cov(x_{m_1},x_{m_2}) - \gd(m_1,m_2,m_3)\ebb(x_{m_1}) \notag \\
 &\quad + \gd(m_1,m_2) \cov(x_{m_1},x_{m_3}) - \gd(m_1,m_2,m_3)\ebb(x_{m_1})  \notag \\
 &= \cum(y_{m_1},y_{m_2},y_{m_3}) - 2\gd(m_1,m_2,m_3)\ebb(x_{m_1})   \notag \\
 &\quad + \gd(m_2,m_3) \cov(x_{m_1},x_{m_2}) 
 + \gd(m_1,m_3) \cov(x_{m_1},x_{m_2}) 
 + \gd(m_1,m_2) \cov(x_{m_1},x_{m_3})  \notag \\
 &= \sbra{\cum(\ga,\ga,\ga)(D^{\top},D^{\top},D^{\top})}_{m_1m_2m_3} - 2\gd(m_1,m_2,m_3) \ebb(x_{m_1})   \notag \\
 &\quad + \gd(m_2,m_3) \cov(x_{m_1},x_{m_2}) 
 + \gd(m_1,m_3) \cov(x_{m_1},x_{m_2}) 
 + \gd(m_1,m_2) \cov(x_{m_1},x_{m_3}),
 \label{cum3}
\end{align}
where, in the third equality, we used the previous result from~\eqref{cum2} that $\cov(y,y) = \cov(x,x) - \diag(\ebb(x))$.

\subsubsection{The diagonal structure of the GP/DICA cumulants} \label{sec:app:dicacum2}
In this section, we provide detailed derivation of the diagonal structure~\eqref{diagS} of the matrix $S$~\eqref{S} and the diagonal structure~\eqref{diagT} of the tensor $T$~\eqref{T}. 

From the independence of $\ga_{1},\ga_{2},\dots,\ga_{K}$ and by the independence property of cumulants from Section~\ref{sec:cumulants},
it follows that  $\cov(\ga,\ga)$ is a diagonal matrix and $\cum(\ga,\ga,\ga)$ is a diagonal tensor, i.e., $\cov(\ga_{k_1},\ga_{k_2})=\gd(k_1,k_2)\cov(\ga_{k_1},\ga_{k_2})$ and 
 $\cum(\ga_{k_1},\ga_{k_2},\ga_{k_3})=\gd(k_1,k_2,k_3) \cum(\ga_{k_1},\ga_{k_1},\ga_{k_1})$. Therefore, the following holds
\begin{ma}
\cov(y_{m_1},y_{m_2}) &= \sum_k  \cov(\ga_k,\ga_k) D_{m_1k}D_{m_2k}, \\
\cum(y_{m_1},y_{m_2},y_{m_3}) &= \sum_k \cum(\ga_k,\ga_k,\ga_k) D_{m_1k}D_{m_2k}D_{m_3k}, 
\end{ma}
which we can rewrite in a matrix/tensor form as
\begin{ma}
\cov(y,y) &= \sum_k \cov(\ga_k,\ga_k) d_k d_k^{\top}, \\
\cum(y,y,y) &= \sum_k \cum(\ga_k,\ga_k,\ga_k) d_k\tp d_k\tp d_k.
\end{ma}
 Moving $\cov(y,y)$ / $\cum(y,y,y)$ in the expression for $\cov(x,x)$~\eqref{cum2} / $\cum(x,x,x)$~\eqref{cum3} on one side of equality and all other terms on the other side, we define matrix $S\in\rr{M\times M}$  / tensor $T\in\rr{M\times M\times M}$ as follows
\begin{align}
\label{appSgp} S &:=\cov(x,x) - \diag\rbra{\ebb(x)}, \\
\nonumber T_{m_1m_2m_3} &:=  \cum(x_{m_1},x_{m_2},x_{m_3})+ 2\gd(m_1,m_2,m_3)\ebb(x_{m_1}) \\
\nonumber &- \gd(m_2,m_3)\cov(x_{m_1},x_{m_2})\\
\nonumber & - \gd(m_1,m_3)\cov(x_{m_1},x_{m_2})\\
\label{appTgp}& - \gd(m_1,m_2) \cov(x_{m_1},x_{m_3}).
\end{align}
By construction, $S=\cov(y,y)$ and $T=\cum(y,y,y)$ and, therefore, it holds that
\begin{align}
\label{appSdiaggp} S =& \sum_k \cov(\ga_k,\ga_k) d_k d_k^{\top}, \\
\label{appTdiaggp} T =& \sum_k \cum(\ga_k,\ga_k,\ga_k) d_k\tp d_k\tp d_k.
\end{align}
This means that both the matrix $S$ and the tensor $T$ are sums of rank-1 matrices and tensors, respectively\footnote{For tensors, such decomposition is also known under the names CANDECOMP/PARAFAC or, simply, the CP decomposition (see, e.g.,~\citesup{KolBad2009}). }. This structure of the matrix $S$ and the tensor $T$ is the basis for the algorithms considered in this paper.

\subsubsection{Unbiased finite sample estimators for the GP/DICA cumulants} \label{sec:dica:fs}
Given a sample $\cbra{x_1,x_2,\dots,x_N}$, we obtain a finite sample estimate $\wh{S}$ of $S$~\eqref{S} / $\wh{T}$ of $T$~\eqref{T} for the GP/DICA cumulants:
\begin{align}
\label{appSgp:fs} \wh{S} &:=\wcov(x,x) - \diag\rbra{\webb(x)}, \\
\nonumber \wh{T}_{m_1m_2m_3} &:=  \wcum(x_{m_1},x_{m_2},x_{m_3})+ 2\gd(m_1,m_2,m_3)\webb(x_{m_1}) \\
\nonumber &- \gd(m_2,m_3)\wcov(x_{m_1},x_{m_2})\\
\nonumber & - \gd(m_1,m_3)\wcov(x_{m_1},x_{m_2})\\
\label{appTgp:fs}& - \gd(m_1,m_2) \wcov(x_{m_1},x_{m_3}),
\end{align}
where unbiased estimators of the first three cumulants are
\begin{equation}\label{cum:empirical}
\begin{aligned}
\wh{\ebb}(x_{m_1}) =& \frac{1}{N} \sum_n x_{nm_1}, \\
\wh{\cov}(x_{m_1},x_{m_2}) =& \frac{1}{N-1} \sum_n z_{nm_1}  z_{nm_2}, \\
\wh{\cum}(x_{m_1},x_{m_2},x_{m_3}) =& \frac{N}{(N-1)(N-2)} \sum_n z_{nm_1}  z_{nm_2}  z_{nm_3},
\end{aligned}
\end{equation}
where the word vocabulary indexes are $m_1, m_2, m_3 = 1,2,\dots,M$ and the centered documents $z_{nm} := x_{nm} - \wh{\ebb}(x_m)$. (The latter is introduced only for compact representation of~\eqref{cum:empirical} and is different from $z$ in the LDA model.)

\subsubsection{On the orders of cumulants}\label{sec:on-the-orders-of-cumulants}
Note that the factorization of $S = \wt{D}\wt{D}^{\top}$ does not uniquely determine $\wt{D}$ as one can equivalently use $S = (\wt{D}U)(\wt{D}U)^{\top}$ with any orthogonal $K \times K$ matrix $U$. Therefore, one has to consider higher than the second order information. Moreover, in ICA the fourth-order tensors are used, because the third cumulant of the Gaussian distribution is zero, which is not the case in the DICA/LDA models, where the third order information is sufficient.

\subsection{Appendix. The sketch of the proof for  Proposition~\ref{sample-complexity}} \label{sec:sample-complexity}
\subsubsection{Expected squared error for the sample expectation} \label{section-variance-expectation}
The sample expectation is $\webb(x) = \frac{1}{N} \sum_n x_n$ is an unbiased estimator of the expectation and:
\begin{ma}
\ebb\rbra{\normp{\webb(x) - \ebb(x)}_2^2} & = \sum_m \ebb\sbra{\rbra{\webb(x_m) - \ebb(x_m)}^2} \\
& = \frac{1}{N^2} \sum_m \sbra{ \ebb\rbra{ \sum_n \rbra{x_{nm} - \ebb(x_m)}^2 } + \ebb\rbra{ \sum_n \sum_{n\ne n'} \rbra{x_{nm} - \ebb(x_m)} \rbra{x_{n'm} - \ebb(x_m)}} } \\
& = \frac{1}{N} \sum_m \ebb\sbra{ \rbra{x_{m} - \ebb(x_m)}^2 } = \frac{1}{N}\sum_m \var(x_m).
\end{ma}
Further, by the law of total variance:
\begin{ma}
\ebb\rbra{\normp{\webb(x) - \ebb(x)}_2^2} &= \frac{1}{N}\sum_m \sbra{ \ebb(\var(x_m|y)) + \var(\ebb(x_m|y)) } = \frac{1}{N} \sum_m \sbra{ \ebb(y_m) + \var(y_m) } \\
& = \frac{1}{N}\sbra{ \sum_k \ebb(\ga_k) + \sum_k \innerp{d_k,d_k} \var(\ga_k)},
\end{ma}
using the fact that $\sum_m D_{mk} = 1$ for any $k$.

\subsubsection{Expected squared error for the sample covariance} \label{section-variance-covariance}
The following finite sample estimator of the covariance $\cov(x,x) = \ebb(xx^{\top}) - \ebb(x)\ebb(x)^{\top}$
\begin{equation}\label{sample-covariance}
\begin{aligned}
\wcov(x,x) &= \frac{1}{N-1} \sum_n x_nx_n^{\top} - \webb(x)\webb(x)^{\top} 
 = \frac{1}{N-1} \sum_n \rbra{x_n x_n^{\top} - \frac{1}{N^2}  \sum_{n'}\sum_{n''} x_{n'}x_{n''}^{\top} }
\\
&= \frac{1}{N}\sum_n\rbra{x_nx_n^{\top} - \frac{1}{N-1}x_n\sum_{n'\ne n} x_{n'}^{\top} }
\end{aligned}
\end{equation}
is unbiased, i.e., $\ebb(\wcov(x,x)) = \cov(x,x)$. Its squared error is
\begin{ma}
&\ebb\rbra{\normp{\wcov(x,x) - \cov(x,x)}_F^2} = \sum_{m,m'}
\ebb\sbra{ \rbra{\wcov(x_m,x_{m'}) - \ebb[\wcov(x_m,x_{m'})] }^2 }.
\end{ma}
The $m,m'$-th element of the sum above is equal to
\begin{ma}
& \frac{1}{N^2}  \sum_{n,n'} \cov\rbra{ x_{nm}x_{nm'} - \frac{1}{N-1}x_{nm}\sum_{n''\ne n}x_{n''m'}\quad , \quad x_{n'm} x_{n'm'} - \frac{1}{N-1}x_{n'm}\sum_{n'''\ne n'}x_{n'''m'}} \\
&=\frac{1}{N^2}  \sum_{n,n'} \cov\rbra{x_{nm}x_{nm'}, x_{n'm} x_{n'm'}}
-\frac{2}{N^2(N-1)} \sum_{n,n'} \cov\rbra{x_{nm}\sum_{n''\ne n} x_{n''m'}, x_{n'm}x_{n'm'}}\\
& + \frac{1}{N^2(N-1)^2} \sum_{n,n'} \cov\rbra{x_{nm}\sum_{n''\ne n} x_{n''m'},x_{n'm}\sum_{n'''\ne n'} x_{n'''m'} }\\
& =\frac{1}{N^2} \sum_n \cov\rbra{x_{nm}x_{nm'},x_{nm}x_{nm'}} \\
& - \frac{2}{N^2(N-1)} \sbra{ \sum_n\sum_{n''\ne n} \cov\rbra{x_{nm}x_{n''m'},x_{nm}x_{nm'}} 
 +  \sum_n\sum_{n'\ne n}  \cov\rbra{ x_{nm}x_{n'm'},x_{n'm}x_{n'm'}}  } \\ 
& + \frac{1}{N^2(N-1)^2} \sbra{  \sum_n \sum_{n'' \ne n} \sum_{n''' \ne n} \cov\rbra{x_{nm}x_{n''m'},x_{nm}x_{n'''m'}}  +   \sum_{n'} \sum_{n\ne n'} \sum_{n'' \ne n} \cov\rbra{ x_{nm}x_{n''m'},x_{n'm}x_{nm'}} } \\
& + \frac{1}{N^2(N-1)^2} \sbra{ \sum_{n'} \sum_{n\ne n'} \sum_{n''' \ne n'} \cov\rbra{ x_{nm}x_{n'm'},x_{n'm}x_{n'''m'}}  +  \sum_{n'} \sum_{n\ne n'} \sum_{n'' \ne n} \cov\rbra{ x_{nm}x_{n''m'},x_{n'm}x_{n''m'}} },
\end{ma}
where we used mutual independence of the observations $x_n$ in a sample $\cbra{x_n}_{n=1}^N$ to conclude that the covariance between the two expressions involving only independent variables is zero.

Further:
\begin{ma}
&\ebb\rbra{\normp{\wcov(x,x) - \cov(x,x)}_F^2} = \frac{1}{N^2} \sum_{m,m'} N \rbra{ \ebb(x_m^2x_{m'}^2) - \sbra{\ebb(x_mx_{m'})}^2 } \\
& - \frac{4}{N^2(N-1)} \sum_{m,m'} N(N-1) \rbra{ \ebb(x_m^2x_{m'})\ebb(x_{m'}) - \ebb(x_mx_{m'})\ebb(x_m)\ebb(x_{m'}) } \\
& + \frac{2}{N^2(N-1)^2} \sum_{m,m'} N (N-1) (N-2) \rbra{\ebb(x_m^2)\sbra{\ebb(x_{m'})}^2 - \sbra{\ebb(x_m)}^2 \sbra{\ebb(x_{m'})}^2 } \\
& + \frac{2}{N^2(N-1)^2} \sum_{m,m'} N(N-1)(N-2) \rbra{\ebb(x_mx_{m'})\ebb(x_{m})\ebb(x_{m'}) - \sbra{\ebb(x_m)}^2 \sbra{\ebb(x_{m'})}^2 } + O\rbra{\frac{1}{N^2} },
\end{ma}
which after simplification gives
\begin{ma}
&\ebb\rbra{\normp{\wcov(x,x) - \cov(x,x)}_F^2} = \frac{1}{N} \sum_{m,m'} \sbra{ \var(x_mx_{m'}) + 2 \sbra{\ebb(x_m)}^2\var(x_{m'}) } \\
& + \frac{1}{N}\sum_{m,m'} \sbra{ 2 \ebb(x_m)\ebb(x_{m'}) \cov(x_m,x_{m'}) - 4 \ebb(x_m)\cov(x_mx_{m'},x_{m'}) } + O\rbra{\frac{1}{N^2}},
\end{ma}
where in the last equality, by symmetry, the  summation indexes $m$ and $m'$ can be exchanged.
As $x_m \sim \poi(y_m)$, by the law of total expectation and law of total covariance, it follows, for $m \neq m'$ (and using the auxiliary expressions from Section~\ref{auxiliary-expressions-fsc}):
\begin{ma}
\var(x_mx_{m'}) & =  \ebb(x_m^2x_{m'}^2) - \sbra{\ebb[x_mx_{m'}]}^2 = \ebb\sbra{\ebb(x_m^2x_{m'}^2|y)} - \sbra{ \ebb\sbra{ \ebb(x_mx_{m'}|y) } }^2 \\
& = \ebb\sbra{ y_m^2 y_{m'}^2 + y_m^2 y_{m'} + y_my_{m'}^2 + y_my_{m'}  }   - \sbra{ \ebb(y_my_{m'}) }^2, \\
\sbra{\ebb(x_m)}^2\var(x_{m'}) & = \sbra{\ebb(y_m)}^2 \ebb(y_{m'}) + \sbra{\ebb(y_m)}^2\ebb(y_{m'}^2) - \sbra{\ebb(y_m)}^2\sbra{\ebb(y_{m'})}^2, \\
\ebb(x_m)\ebb(x_{m'}) \cov(x_m,x_{m'}) & =  \ebb(y_my_{m'}) \ebb(y_m)\ebb(y_{m'}) - \sbra{ \ebb(y_m) }^2 \sbra{ \ebb(y_{m'}) }^2, \\
 \ebb(x_m) \cov(x_mx_{m'},x_{m'}) & =  \ebb(y_m) \sbra{ \ebb(y_my_{m'}) + \ebb(y_my_{m'}^2) - \ebb(y_my_{m'}) \ebb(y_{m'}) }.
\end{ma}
Now, considering the $m = m'$ case, we have:
\begin{ma}
\var(x_m^2) & = \ebb[\ebb(x_m^4|y)] - \sbra{ \ebb[\ebb(x_m^2|y)] }^2 \\
& = \ebb\sbra{ y_m^4 + 6 y_m^3 +7 y_m^2 + y_m} - \sbra{ \ebb\sbra{ y_m^2 + y_m} }^2, \\
\ebb(x_m)\ebb(x_{m}) \cov(x_m,x_{m}) & =  \ebb(y_m)^2 \sbra{ \ebb(y_m^2) + \ebb(y_m) - \sbra{ \ebb(y_{m}) }^2} , \\
 \ebb(x_m) \cov(x_m^2,x_{m}) & =  \ebb(y_m) \sbra{ \ebb(y_m^3) +  3\ebb(y_m^2)+\ebb(y_m) - \ebb(y_m) \sbra{\ebb(y_m^2)+\ebb(y_m)  } }.
\end{ma}
Substitution of $y_m = \sum_kD_{mk}\ga_k$ gives the following
\begin{ma}
\ebb\rbra{\normp{\wcov(x,x) - \cov(x,x)}_F^2} & = \frac{1}{N}\sum_{k,k',k'',k'''}
\innerp{d_k,d_{k'}}\innerp{d_{k''},d_{k'''}}
 \acal_{kk'k''k'''}
\\
& + \frac{1}{N}\sum_{k,k',k''} \sbra{
\innerp{d_k,d_{k'}}\innerp{d_{k''},\ones}
 \bcal_{kk'k''} + \innerp{d_k\circ d_{k'}, d_{k''}} \ecal_{kk'k''} } \\
 & + \frac{1}{N}\sum_{k,k'} \sbra{ \innerp{d_k,\ones}\innerp{d_{k'},\ones} \ebb(\ga_k\ga_{k'}) + \innerp{d_k,d_{k'}} \fcal_{kk'} } \\
 & + \sum_k \innerp{d_k,\ones} \ebb(\ga_k) + O\rbra{\frac{1}{N^2}},
\end{ma}
where $\ones$ is the vector with all the elements equal to $1$ and
\begin{ma}
\acal_{kk'k''k'''} &=  \ebb(\ga_k\ga_{k'}\ga_{k''}\ga_{k'''}) 
- \ebb(\ga_k\ga_{k''}) \ebb(\ga_{k'}\ga_{k'''}) 
+ 2\ebb(\ga_k)\ebb(\ga_{k'})\ebb(\ga_{k''}\ga_{k'''})\\
&- 2 \ebb(\ga_k)\ebb(\ga_{k'})\ebb(\ga_{k''})\ebb(\ga_{k'''})
+2 \ebb(\ga_k\ga_{k''})\ebb(\ga_{k'})\ebb(\ga_{k'''})  -2\ebb(\ga_k)\ebb(\ga_{k'})\ebb(\ga_{k''})\ebb(\ga_{k'''})\\
&-4\ebb(\ga_k)\ebb(\ga_{k'}\ga_{k''}\ga_{k'''})
+4\ebb(\ga_k)\ebb(\ga_{k'}\ga_{k''})\ebb(\ga_{k'''}),\\
\bcal_{kk'k''} & = 2\ebb(\ga_k\ga_{k'}\ga_{k''}) + 2\ebb(\ga_k)\ebb(\ga_{k'})\ebb(\ga_{k''}) - 4\ebb(\ga_k)\ebb(\ga_{k'}\ga_{k''}), \\
\ecal_{kk'k''} &= 4 \ebb(\ga_k\ga_{k'}\ga_{k''}) + 6 \ebb(\ga_k)\ebb(\ga_{k'}) \ebb(\ga_{k''}) - 10\ebb(\ga_k\ga_{k'})\ebb(\ga_{k''}), \\
\fcal_{kk'} & = 6 \ebb(\ga_k\ga_{k'}) - 5 \ebb(\ga_k)\ebb(\ga_{k'}),
\end{ma}
where we used the expressions from Section~\ref{auxiliary-expressions-fsc}.

\subsubsection{Expected squared error of the  estimator \texorpdfstring{$\mathbf{\wh{S}}$}{S} for the GP/DICA cumulants} \label{varS}
As the estimator $\wh{S}$~\eqref{appSgp:fs} of $S$~\eqref{S} is unbiased, its expected squared error is
\begin{equation}\label{variance-S}
\begin{aligned}
\ebb\sbra{ \normp{\wh{S} - S}_F^2 } =&
  \ebb\sbra{ \norm{ \rbra{\wcov(x,x) - \cov(x,x)} + \rbra{ \diag [\webb(x)] - \diag\sbra{\ebb(x)} } }_F^2 } \\
& = \ebb\sbra{\normp{\webb(x)-\ebb(x)}_F^2} + \ebb\sbra{\normp{\wcov(x,x) - \cov(x,x)}_F^2} \\
& + 2\sum_{m} \ebb\sbra{ \rbra{ \webb(x_m)-\ebb(x_m) }\rbra{ \wcov(x_m,x_{m}) - \cov(x_m,x_{m}) } }.
\end{aligned}
\end{equation}
As $\webb(x_m)$ and $\wcov(x_m,x_m)$ are unbiased,  the $m$-th element of the last sum is equal to
\begin{ma}
&\cov\sbra{ \webb(x_m),\wcov(x_m,x_m) }  \\
&= \frac{1}{N^2} \sum_{n,n'} \cov\sbra{ x_{nm}, x_{n'm}^2} - \frac{1}{N^2(N-1)} \sum_{n,n',n''\ne n'} \cov\sbra{ x_{nm},  x_{n'm}x_{n''m}} \\
&= \frac{1}{N^2} \sum_{n} \cov\sbra{ x_{nm}, x_{nm}^2} - \frac{2}{N^2(N-1)} \sum_{n,n'\ne n} \cov\sbra{ x_{nm},  x_{n'm}x_{nm}} +  O\rbra{\frac{1}{N^2}}\\
& = \frac{1}{N} \ebb(x_m^3) - \frac{2}{N}\left( \ebb(x_m^2)\ebb(x_m) - \sbra{ \ebb(x_m) }^3 \right) + O\rbra{\frac{1}{N^2}} \\
& \leq \frac{1}{N} \ebb(x_m^3) + \frac{2}{N} \sbra{ \ebb(x_m) }^3  + O\rbra{ \frac{1}{N^2} } = \frac{1}{N}\sbra{ \ebb(y_m^3) + 3\ebb(y_m^2) + \ebb(y_m) + 2\sbra{\ebb(y_m)}^3 } + O\rbra{\frac{1}{N^2}},
\end{ma}
where we neglected the negative term $-\ebb(x_m^2)\ebb(x_m)$ for the inequality, and the last equality follows from the expressions in Section~\ref{auxiliary-expressions-fsc}. Further, the fact that $y_m = \sum_k D_{mk}\ga_k$ gives
\begin{ma}
\sum_m\cov\sbra{ \webb(x_m),\wcov(x_m,x_m) } & = \frac{1}{N} \sum_{k,k',k''} \innerp{d_k\circ d_{k'}, d_{k''}} \ccal_{kk'k''} \\
& + \frac{3}{N}\sum_{k,k'} \innerp{d_k,d_{k'}} \ebb(\ga_k\ga_{k'}) + \frac{1}{N} \sum_k \innerp{d_k,\ones} \ebb(\ga_k) + O\rbra{ \frac{1}{N^2} },
\end{ma}
where $\circ$ denotes the elementwise Hadamard product and 
\begin{ma}
\ccal_{kk'k''} & = \ebb(\ga_k\ga_{k'}\ga_{k''}) + 2\ebb(\ga_k)\ebb(\ga_{k'})\ebb(\ga_{k''}).
\end{ma}
Plugging this and the expressions for $\ebb(\normp{\webb(x)-\ebb(x)}_F^2)$  and $\ebb(\normp{\wcov(x,x)-\cov(x,x)}_F^2)$ from Sections~\ref{section-variance-expectation} and~\ref{section-variance-covariance}, respectively, into \eqref{variance-S} gives
\begin{ma}
& \ebb\sbra{ \normp{\wh{S} - S}_F^2 } = 
\frac{1}{N}\sbra{ \sum_k \innerp{d_k,d_k} \var(\ga_k) + \sum_k \ebb(\ga_k)
 +\sum_{k,k',k'',k'''}
\innerp{d_k,d_{k'}}\innerp{d_{k''},d_{k'''}}
 \acal_{kk'k''k'''} } + O\rbra{ \frac{1}{N^2} }\\
& + \frac{1}{N} \sbra{ \sum_{k,k',k''}  \sbra{
\innerp{d_k,d_{k'}}
 \bcal_{kk'k''}  
+ 2 \innerp{d_k\circ d_{k'}, d_{k''}} \ccal_{kk'k''} } +  \sum_{k,k'} \rbra{1 + 6 \inner{d_k,d_{k'}} }\ebb(\ga_k\ga_{k'}) + 2\sum_k  \ebb(\ga_k)},
\end{ma}
where we used that, by the simplex constraint on the topics, $\innerp{d_k,\ones} = 1$ for all $k$. To analyze this expression in more details, let us now consider the GP model, i.e., $\ga_k\sim\gam(c_k,b)$:
\begin{ma}
\sum_{k,k',k'',k'''} \acal_{kk'k''k'''} & \le \frac{30c_0^4 + 23 c_0^3 + 14c_0^2 + 8 c_0}{b^4}, \quad\mbox{and}\quad
\sum_{k,k',k''} \bcal_{kk'k''}  \le \frac{6c_0^3 + 10 c_0^2 + 4c_0}{b^3},
\\
\sum_{k,k',k''} \ccal_{kk'k''} & \le \frac{7c_0^3 + 6c_0^2 + 2c_0}{b^3}, \quad\mbox{and}\quad \sum_{k,k',k''} \ecal_{kk'k''} \le \frac{12c_0^3 + 10c_0^2 + 8c_0}{b^3}, \\
\sum_{k,k'} \fcal_{kk'} & \le \frac{2c_0^2 + c_0}{b^2}\quad\mbox{and}\quad
\sum_{k,k'} \ebb(\ga_k\ga_{k'}) \le \frac{2c_0^2 + c_0}{b^2},
\end{ma}
where we used the expressions from Section~\ref{auxiliary-expressions-fsc}, which gives
\begin{ma}
& \ebb\sbra{ \normp{ \wh{S} - S }_F^2 } \le \frac{\nu}{N} \sbra{  \max_k \normp{d_k}_2^2 \frac{c_0}{b^2} + \frac{c_0}{b} + \rbra{\max_{k,k'} \innerp{d_k,d_{k'}}}^2 \max\sbra{ \frac{c_0^4}{b^4},\frac{c_0}{b^4} } + \max_{k,k'} \innerp{d_k,d_{k'}} \max\sbra{ \frac{c_0^3}{b^3}, \frac{c_0}{b^3} } } \\
& + \frac{\nu}{N} \sbra{   \rbra{ \max_{k,k',k''}\innerp{d_k\circ d_{k'},d_{k''}} } \max\sbra{ \frac{c_0^3}{b^3}, \frac{c_0}{b^3} }  + \rbra{1+\max_{k,k'}\innerp{d_k,d_{k'}} } \max\sbra{\frac{c_0^2}{b^2},\frac{c_0}{b^2}}  } + O\rbra{\frac{1}{N^2}},
\end{ma}
where $\nu\le 30$ is a universal constant.
As, by the Cauchy-Schwarz inequality, $\max\mathop{}_{k,k'} \innerp{d_k,d_{k'}} \le \max\mathop{}_k \norm{d_k}_2^2=:\Delta_1$ and $\max\mathop{}_{k,k',k''} \innerp{d_k\circ d_{k'},d_{k''}} \le \max\mathop{}_k \norm{d_k}_\infty \norm{d_k}_2^2 \leq \max\mathop{}_k \norm{d_k}_2^3 =:\Delta_2$ (note that for the topics in the simplex, $\Delta_2 \leq \Delta_1$ as well as $\Delta_1^2 \le \Delta_1$), it follows that
\begin{ma}
& \ebb\sbra{ \normp{ \wh{S} - S }_F^2 } \le \frac{\nu}{N} \sbra{ \Delta_1\rbra{\frac{L^2}{\bar{c}_0} + \frac{L^3}{\bar{c}_0^2} }  + L + \Delta_1^2\frac{L^4}{\bar{c}_0^3}  + \frac{L^2}{\bar{c}_0^2} + \Delta_2 \frac{L^3}{\bar{c}_0^2} } + O\rbra{\frac{1}{N^2} } \\
&  \le \frac{2\nu}{N} \frac{1}{\bar{c}_0^3} \sbra{ \Delta_1^2L^4 + \bar{c}_0  \Delta_1  L^3   +\bar{c}_0^2L^2  + \bar{c}_0^3 L} + O\rbra{\frac{1}{N^2} },
\end{ma}
where $\bar{c}_0 = \min(1,c_0)\le1$
 and, from
Section~\ref{sec:L:gp}, $c_0 = bL$ where $L$ is the expected document length. The second term $\bar{c}_0\Delta_1L^3$ cannot be dominant as the system $\bar{c}_0\Delta_1L^3>\bar{c}_0^2L^2$ and $\bar{c}_0\Delta_1L^3>\Delta_1^2L^4$ is infeasible. Also, with the reasonable assumption that $L \geq 1$, we also have that the 4th term $\bar{c}_0^3 L \leq \bar{c}_0^2L^2$. 
Therefore,
\begin{ma}
& \ebb\sbra{ \normp{ \wh{S} - S }_F^2 }  \le \frac{3\nu}{N}  \max\sbra{ \Delta_1^2L^4,\,\bar{c}_0^2L^2  } + O\rbra{\frac{1}{N^2} }.
\end{ma}

\subsubsection{Auxiliary expressions}\label{auxiliary-expressions-fsc}
As $\cbra{x_m}_{m=1}^M$ are conditionally independent given $y$ in the DICA model~\eqref{gp}, we have the following expressions by using the law of total expectation for $m \neq m'$ and using the moments of the Poisson distribution with parameter $y_m$:
\begin{ma}
\ebb(x_m) &= \ebb[\ebb(x_m|y_m)] = \ebb( y_m),\\
\ebb(x_m^2) &= \ebb[\ebb(x_m^2|y_m)] =  \ebb(y_m^2) + \ebb(y_m), \\
\ebb(x_m^3) &= \ebb[\ebb(x_m^3|y_m)] = \ebb( y_m^3) + 3\ebb(y_m^2) + \ebb(y_m), \\
\ebb(x_m^4) &= \ebb[\ebb(x_m^4|y_m)] = \ebb( y_m^4) + 6\ebb(y_m^3) + 7\ebb(y_m^2) + \ebb(y_m) , \\
\ebb(x_mx_{m'}) &=\ebb[\ebb(x_mx_{m'}|y)] = \ebb[\ebb(x_m|y_m)\ebb(x_{m'}|y_{m'})] = \ebb(y_my_{m'}), \\
\ebb(x_mx_{m'}^2) &= \ebb[\ebb(x_mx_{m'}^2|y)] = \ebb[\ebb(x_m|y_m)\ebb(x_{m'}^2|y_{m'})] = \ebb(y_my_{m'}^2) + \ebb(y_m y_{m'}), \\
\ebb(x_m^2x_{m'}^2) &= \ebb[\ebb(x_m^2|y_m)\ebb(x_{m'}^2|y_{m'})] = \ebb(y_m^2y_{m'}^2) +\ebb( y_m^2y_{m'}) + \ebb(y_my_{m'}^2) + \ebb(y_my_{m'}).
\end{ma}

Moreover, the moments of $\ga_k\sim\gam(c_k,b)$ are
$$
\ebb(\ga_k) = \frac{c_k}{b},\quad \ebb(\ga_k^2) = \frac{c_k^2 + c_k}{b^2}, \quad \ebb(\ga_k^3) = \frac{c_k^3 + 3 c_k^2 + 2 c_k}{b^3}, \quad \ebb(\ga_k^4) = \frac{c_k^4 + 6c_k^3 + 11c_k^2 + 6c_k}{b^4},\quad\mbox{etc.}
$$

\subsubsection{Analysis of the whitening and recovery error}\label{section-analysis-of-whitening}
We can follow a similar analysis as in Appendix C of~\cite{AnaEtAl2013} to derive the topic recovery error given the sample estimate error. In particular, if we define the following sampling errors $E_S$ and $E_T$:
\begin{ma}
&\normp{\wh{S} - S}  \le E_S,\\
&\normp{\wh{T}(u) - T(u)} \le \norm{u}_2E_T,
\end{ma}
then the following form of their Lemma~C.2 holds for both the LDA moments and the GP/DICA cumulants: 
\begin{equation}
\label{analysis-whitening-bound}
\begin{aligned}
\normp{\wh{W} \wh{T}(\wh{W}^{\top}u) \wh{W}^{\top} - WT(W^{\top} u)W^{\top}} &\le \nu\sbra{  \frac{(\mathrm{max}_k \gamma_k) E_S }{\sigma_K \big(\wt{D}\big)^2} + \frac{E_T}{\sigma_K\big(\wt{D}\big)^3} },
\end{aligned}
\end{equation}
where $\sigma_k(\cdot)$ denotes the $k$-th singular value of a matrix, $\nu$ is some universal constant, and in both cases $\wt{D}$ was defined such that $S = \wt{D} \wt{D}^\top$. For the LDA moments, $\gamma_k = 2 \sqrt{ \frac{c_0 (c_0+1))}{c_k (c_0+2)^2 }}$, whereas for the GP/DICA cumulants, $\gamma_k$ takes the simpler form $\gamma_k := \cum(\ga_k)/[\var(\ga_k)]^{3/2}=2/\sqrt{c_k}$.

We note that the scaling for $S$ is $O(L^2)$ for the GP/DICA cumulants, in contrast to $O(1)$ for the LDA moments. Thus, to compare the upper bound~\eqref{analysis-whitening-bound} for the two types of moments, we need to put it in quantities which are common. In the first section of the Appendix~C of~\cite{AnaEtAl2013}, it was mentioned that  $\sigma_K\big(\wt{D}\big) \geq \sqrt{\frac{c_{\mathrm{min}}}{c_0 (c_0+1)}} \sigma_K(D)$ for the LDA moments, where $c_{\mathrm{min}} := \min_k c_k$. In contrast, for the GP/DICA cumulants, we can show that $\sigma_K\big(\wt{D}\big) \geq L \frac{\sqrt{c_{\mathrm{min}}}}{c_0} \sigma_K(D)$, where $L := c_0/b$ is the average length of a document in the GP model. Using this lower bound for the singular vector, we thus get the following bound in the case of the GP cumulant:
\begin{equation} \label{eq:W_bound_us}
\normp{\wh{W} \wh{T}(\wh{W}^{\top}u) \wh{W}^{\top} - WT(W^{\top} u)W^{\top}} 
 \le \frac{\nu}{c_{\mathrm{min}}^{3/2}}\sbra{\frac{E_S}{L^2} \frac{ 2c_0^2 }{\sbra{ \sigma_K\big(D\big) }^2 } + \frac{E_T}{L^3} \frac{c_0^3}{ { \sbra{\sigma_K(D)}^3 } } }.
\end{equation}
The $c_{\mathrm{min}}^{3/2}$ factor is common for both the LDA moment and GP cumulant, but as we mentioned after Proposition~\ref{sample-complexity}, the sample error $E_S$ term gets divided by $L^2$ for the GP cumulant, as expected.

The recovery error bound in~\cite{AnaEtAl2013} is based on the bound~\eqref{eq:W_bound_us}, and thus by showing that the error $E_S/L^2$ for the GP cumulant is lower than the $E_S$ term for the LDA moment, we expect to also gain a similar gain for the recovery error, as the rest of the argument is the same for both types of moments (see Appendix C.2, C.3 and~C.4 in~\cite{AnaEtAl2013} for the completion).

\subsection{Appendix. The LDA moments}\label{sec:lda:moms}
\subsubsection{Our notation} \label{sec:lda:moms:notation}
The LDA moments were derived in~\cite{AnaEtAl2012}. Note that the full version of the paper with proofs appeared in~\cite{AnaEtAl2013} and a later version of the paper also appeared in~\citesup{AnaEtAl2014b}. In this section, we recall the form of the LDA moments using our notation. This section does not contain any novel results and is included for the reader's convenience. We also refer to this section when deriving the practical expressions for computation of the sample estimates of the LDA moments in Appendix~\ref{sec:lda:empirical}.

For deriving the LDA moments, a document is assumed to be composed of at least three tokens: $L \geq 3$. As the LDA generative model~\eqref{lda-tokens} is only defined \emph{conditional} on the length $L$, this is not too problematic. But given that we present models in this paper which also model $L$, we mention for clarity that we can suppose that all expectations and probabilities defined below are implicitly conditioning on $L \geq 3$.\footnote{Note that another advantage of the DICA cumulants from Section~\ref{sec:cum} is that they do not require such a somewhat artificial condition: they are well-defined for any document length (even a document of length zero!).} The theoretical LDA moments are derived only using the first three words $w_1$, $w_2$ and $w_3$ of a document. But note that since the words $w_{\ell}$'s are conditionally i.i.d. given $\theta$ (for $1 \leq \ell \leq L$), we have $M_3 := \ebb(w_{1}\tp w_{2}\tp w_{3}) = \ebb(w_{\ell_1}\tp w_{\ell_2}\tp w_{\ell_3})$ for any three distinct tokens $\ell_1$, $\ell_2$ and $\ell_3$. The tensor $M_3$ is thus symmetric, and could have been defined using any distinct $\ell_1$, $\ell_2$ and $\ell_3$ that are less than $L$. To highlight this arbitrary choice and to make the links with the U-statistics estimator presented later, we thus use generic distinct $\ell_1$, $\ell_2$ and $\ell_3$ in the definition of the LDA moments below, instead of $\ell_1 = 1$, $\ell_2=2$ and $\ell_3 =1$ as in~\cite{AnaEtAl2012}. 

Using this notation, then by the law of total expectation and the properties of the Dirichlet distribution, the non-central moments\footnote{Note, the difference in the notation for the LDA moments in papers~\cite{AnaEtAl2012} and~\cite{AnaEtAl2014}. In~\cite{AnaEtAl2012}, $M_1=\ebb(w_{\ell_1})$, $M_2=\ebb(w_{\ell_1}\tp w_{\ell_2})$, and $M_3 = \ebb(w_{\ell_1}\tp w_{\ell_2} \tp w_{\ell_3})$. However, in~\cite{AnaEtAl2014},  $M_2$ is equivalent to $S$ in our notation and to $Pairs$ in the notation of~\cite{AnaEtAl2012}; similarly, $M_3$ is $T$ in our notation or $Triples$ in the notation of~\cite{AnaEtAl2012}.}
of the LDA model~\eqref{lda-tokens} take the form~\cite{AnaEtAl2012}:
\begin{align}
\label{appM1} 
M_1 &=\ebb(w_{\ell_1}) = D\frac{c}{c_0},\\
\label{appM2}
M_2 &= \ebb(w_{\ell_1}w_{\ell_2}^{\top})=  \frac{c_0}{c_0+1} M_1  M_1^{\top} + \frac{1}{c_0(c_0+1)}D\diag\rbra{c}D^{\top}, \\
\nonumber
M_3 &=\ebb(w_{\ell_1}\tp w_{\ell_2}\tp w_{\ell_3}) \\ 
\nonumber
&= \frac{c_0}{c_0+2} \sbra{\ebb(w_{\ell_1}\tp w_{\ell_2} \tp M_1) + \ebb(w_{\ell_1} \tp M_1 \tp w_{\ell_3}) + \ebb(M_1\tp w_{\ell_2} \tp w_{\ell_3})}, \\
\label{appM3}
&-\frac{2c_0^3}{c_0(c_0+1)(c_0+2)} M_1\tp M_1 \tp M_1 + \frac{2}{c_0(c_0+1)(c_0+2)}\sumk c_k d_k\tp d_k \tp d_k.
\end{align}
where $\tp$ denotes the tensor product.

Similarly to the GP/DICA cumulants (as discussed in Appendix~\ref{sec:app:dicacum2}), moving the terms in the non-central moments~\eqref{appM1},~\eqref{appM2},~\eqref{appM3}, the following quantities are defined
\begin{align}
\label{S:lda}
(Pairs) = S &:= M_2 - \frac{c_0}{c_0+1} M_1  M_1^{\top}, \hspace{5.3cm} \text{LDA S-moment}\\
\nonumber
(Triples) = T & := M_3 - \frac{c_0}{c_0+2} \sbra{\ebb(w_{\ell_1}\tp w_{\ell_2} \tp M_1) + \ebb(w_{\ell_1} \tp M_1 \tp w_{\ell_3}) + \ebb(M_1\tp w_{\ell_2} \tp w_{\ell_3})} \\
\label{T:lda}
& + \frac{2c_0^2}{(c_0+1)(c_0+2)} M_1\tp M_1 \tp M_1. \hspace{3.7cm} \text{LDA T-moment}
\end{align}
Slightly abusing terminology, we refer to the entities $S$ and $T$ as the ``LDA moments''. 
They have the following diagonal structure
\begin{align}
\label{diagS:lda} S &=  \frac{1}{c_0(c_0+1)} \sumk c_k d_kd_k^{\top},  \\
\label{diagT:lda} T &= \frac{2}{c_0(c_0+1)(c_0+2)} \sumk c_k d_k\tp d_k\tp d_k. 
\end{align}
Note however that this form of the LDA moments has a slightly different nature than the similar form~\eqref{diagS} and~\eqref{diagT} of the GP/DICA cumulants. Indeed, the former is the result of properties of the Dirichlet distribution, while the latter is the result of the independence of $\ga$'s. However, one can think of the elements of a Dirichlet random vector as being almost independent (as, e.g., a Dirichlet random vector can be obtained from independent gamma variables through dividing each by their sum). Also, this closeness of the structures of the LDA moments and the GP cumulants can be explained by the closeness of the respective models as discussed in Section~\ref{sec:2}.

\subsubsection{Asymptotically unbiased finite sample estimators for the LDA moments} \label{sec:lda:fs}
Given realizations $w_{n\ell}$, $n=1,\dots,N$, $\ell=1,\dots,L_n$, of the token random variable $w_{\ell}$, we now give the expressions for the finite sample estimates of $S$~\eqref{S:lda} and $T$~\eqref{T:lda} for the LDA model (and we rewrite them as a function of the sample counts $x_n$).\footnote{Note that because non-linear functions of $\wh{M}_1$ appear in the expression for~$\wh{S}$~\eqref{S:lda:fs} and~$\wh{T}$~\eqref{T:lda:fs}, the estimator is biased, i.e., $\ebb(\wh{S}) \neq S$. The bias is small though: $\normp{\ebb(\wh{S}) -  S} = O(1/N)$ and the estimator is asymptotically unbiased. This is in contrast with the estimator for the GP/DICA moments which is easily made unbiased.} We use the notation $\wh{\ebb}$ below to express a U-statistics empirical expectation over the token within a documents, uniformly averaged over the whole corpus. For example, $\wh{\ebb}(w_{\ell_1}\tp w_{\ell_2} \tp \wh{M}_1) := \frac{1}{N} \sumn \frac{1}{L_n(L_n-1)}\sum_{\ell_1=1}^{L_n} \sum_{\substack{\ell_2=1\\ \ell_2\ne \ell_1}}^{L_n} w_{\ell_1}\tp w_{\ell_2}\tp \wh{M}_1 $.
\begin{align}
\label{S:lda:fs}
\wh{S} &:= \wh{M}_2 - \frac{c_0}{c_0+1} \wh{M}_1  \wh{M}_1^{\top}, \\
\nonumber
\wh{T} & := \wh{M}_3 - \frac{c_0}{c_0+2} \sbra{\wh{\ebb}(w_{\ell_1}\tp w_{\ell_2} \tp \wh{M}_1) + \wh{\ebb}(w_{\ell_1} \tp \wh{M}_1 \tp w_{\ell_3}) + \wh{\ebb}(\wh{M}_1\tp w_{\ell_2} \tp w_{\ell_3})} \\
\label{T:lda:fs}
& + \frac{2c_0^2}{(c_0+1)(c_0+2)} \wh{M}_1\tp \wh{M}_1 \tp \wh{M}_1,
\end{align}
where, as suggested in~\cite{AnaEtAl2014}, unbiased U-statistics estimates of 
$M_1$, $M_2$ and $M_3$ are:
\begin{align}
\label{m1final} \wh{M}_1 & :=\wh{\ebb}(w_{\ell})= \frac{1}{N} \sumn \frac{1}{L_n} \sum_{\ell=1}^{L_n} w_{n\ell} = \frac{1}{N}\sumn [\delta_1]_n x_n  = \frac{1}{N} X\delta_1, \\
\nonumber \wh{M}_2 &:=\wh{\ebb}(w_{\ell_1} w_{\ell_2}^{\top}) = \frac{1}{N} \sumn \frac{1}{L_n(L_n-1)}\sum_{\ell_1=1}^{L_n} \sum_{\substack{\ell_2=1\\ \ell_2\ne \ell_1}}^{L_n}w_{n\ell_1} w_{n\ell_2}^{\top} \\
\nonumber & = \frac{1}{N} \sumn [\delta_2]_n \rbra{x_n x_n^{\top} - \sum_{\ell=1}^{L_n} w_{n\ell} w_{n\ell}^{\top}} \\
\nonumber & = \frac{1}{N} \sumn [\delta_2]_n \rbra{x_n x_n^{\top} - \diag(x_n) } \\ 
\label{m2final} &= \frac{1}{N} \sbra{X \diag(\delta_2)X^{\top} - \diag(X\delta_2)}, \\ \end{align}
\begin{align}
\nonumber
\wh{M}_3&:=\wh{\ebb}(w_{\ell_1}\tp w_{\ell_2}\tp w_{\ell_3}) = \frac{1}{N} \sumn \delta_{3n} \sum_{\ell_1=1}^{L_n} \sum_{\substack{\ell_2=1\\\ell_2\ne \ell_1}}^{L_n}\sum_{\substack{\ell_3=1\\\ell_3\ne\ell_2\\ \ell_3\ne\ell_1}}^{L_n} w_{n\ell_1}\tp w_{n\ell_2}\tp w_{n\ell_3} \\
\nonumber
&= \frac{1}{N}\sumn [\delta_3]_n \left( x_n \tp x_n\tp x_n -\sum_{\ell=1}^{L_n} w_{n\ell}\tp w_{n\ell} \tp w_{n\ell} \right. \\
\nonumber
& \left. - \sum_{\ell_1=1}^{L_n}\sum_{\substack{\ell_2=1\\ \ell_2\ne\ell_1}}^{L_n}(w_{n\ell_1}\tp w_{n\ell_1}\tp w_{n\ell_2} + w_{n\ell_1}\tp w_{n\ell_2}\tp w_{n\ell_1} + w_{n\ell_1}\tp w_{n\ell_2}\tp w_{n\ell_2}) \right) \\
\nonumber
&= \frac{1}{N}\sumn [\delta_3]_n \left( x_n\tp x_n \tp x_n + 2\summ x_{nm} (e_m\tp e_m\tp e_m) \right. \\
\label{m3final}
& \left. - \sum_{m_1=1}^M\sum_{m_2=1}^M x_{nm_1} x_{nm_2} (e_{m_1}\tp e_{m_1} \tp e_{m_2} + e_{m_1}\tp e_{m_2}\tp e_{m_1} + e_{m_1}\tp e_{m_2} \tp e_{m_2}) \right).
\end{align}
Here, the vectors $\gd_1$, $\gd_2$ and $\gd_3 \in\rr{N}$ are defined as $\sbra{\delta_1}_n := L_n^{-1}$; $\sbra{\delta_2}_n := (L_n(L_n-1))^{-1}$, i.e., $\sbra{\delta_2}_n = \sbra{{L_n \choose 2} 2!}^{-1}$ is the number of times to choose an ordered pair of tokens out of $L_n$ 
tokens;  $[\delta_3]_n := (L_n(L_n-1)(L_n-2))^{-1}$, i.e., $[\delta_3]_n = \sbra{{L_n \choose 3} 3!}^{-1}$ is the number of times to choose an ordered triple of tokens out of $L_n$ tokens. Note that the vectors $\gd_1$, $\gd_2$, and $\gd_3$ have nothing to do with the Kronecker delta $\gd$.

For a vector $a\in\rr{N}$, we sometimes use notation $\sbra{a}_n$ to denote its $n$-th element. Similarly, for a matrix $A\in\rr{M\times N}$ we use notation $\sbra{A}_{mn}$ to denote its $(m,n)$-th element.

There is a slight abuse of notation in the expressions above as $w_{\ell}$ is sometimes treated as a random variable (i.e., in $\webb(w_{\ell})$, $\webb(w_{\ell_1} w_{\ell_2}^{\top})$, etc.) and sometimes as its realization. However, the difference is clear from the context.

\subsection{Appendix. Practical aspects and implementation details} \label{sec:app:implementation}
\subsubsection{Whitening of $\mathbf{S}$ and dimensionality reduction}\label{sec:whitening}
The algorithms from Section~\ref{sec:diag} require the computation of a whitening matrix $W$ of $S$. Due to the similar diagonal structure (\eqref{diagS:lda} and~\eqref{diagS}) of the matrix S for both the LDA moments~\eqref{S:lda} and the GP/DICA cumulants~\eqref{S}, the computation of a whitening matrix is exactly the same in both cases. 

By a whitening matrix, we mean a matrix $W\in\rr{K\times M}$ (in practice, $M\gg K$) that does not only whiten $S\in\rr{M\times M}$, but also reduces its dimensionality such that\footnote{Note that such a whitening matrix $W\in\rr{K\times M}$ is not uniquely defined as left multiplication by any orthogonal matrix $V\in\rr{K\times K}$ does not change anything. Indeed, let $\wt{W} = VW$, then $\wt{W}S\wt{W}^{\top} = VWSW^{\top} V^{\top} = I_K$.} $WSW^{\top} = I_K$.

Let $S = U \Sigma U^{\top}$ be an orthogonal eigendecomposition of the symmetric matrix $S$. Let $\Sigma_{1:K}$ denotes the diagonal matrix that contains the largest $K$ eigenvalues\footnote{We mean the largest non-negative eigenvalues. In theory, $S$ have to be PSD. In practice, when we deal with finite number of samples, respective estimate of $S$ can have negative eigenvalues. However, for $K$ sufficiently small, $S$ should have enough positive eigenvalues. Moreover, it is standard practice to use eigenvalues of $S$ for estimation of a good value of $K$, e.g., by thresholding all negative and close to zero eigenvalues.}  
of $S$ on its diagonal and let $U_{1:K}$ be a matrix with the respective eigenvalues in its columns. Then, a whitening matrix is
\begin{equation}\label{W}
W = \Sigma_{1:K}^{\dagger 1/2}U_{1:K}^{\top} ,
\end{equation}
where $\Sigma_{1:K}^{\dagger 1/2}$ is a diagonal matrix constructed from $\Sigma_{1:K}$ by taking the inverse and the square root of its non-zero diagonal values ($^{\dagger}$ stands for the pseudo-inverse).

In practice, when only a finite sample estimator $\wh{S}$ of $S$ is available, the following finite sample estimator $\wh{W}$ of $W$ can be introduced
\begin{equation}\label{W-fs}
\wh{W} :=  \wh{\Sigma}_{1:K}^{\dagger 1/2} \wh{U}_{1:K}^{\top},
\end{equation}
where $\wh{S} = \wh{U} \wh{\Sigma} \wh{U}^{\top}$.

\subsubsection{Computation of the finite sample estimators of the GP/DICA cumulants} \label{sec:dica:empirical}
In this section, we present efficient formulas for computation of the finite sample estimate (see Appendix~\ref{sec:dica:fs} for the definition of $\wh{T}$) of  $\wh{W}\wh{T}(v)\wh{W}^{\top}$ for the GP/DICA models. The construction of the finite sample estimator $\wh{W}$ is discussed in Appendix~\ref{sec:whitening}, while the computation of $\wh{S}$~\eqref{appSgp:fs} is straightforward.

By plugging the definition of the tensor $\wh{T}$~\eqref{appTgp:fs} in the formula~\eqref{projT} for the projection of a tensor onto a vector, we obtain for a given $v\in\rr{M}$:
\begin{ma}
\sbra{\wh{T}(v)}_{m_1m_2} &= \sum_{m_3} \wcum(x_{m_1},x_{m_2},x_{m_3}) v_{m_3} + 2\sum_{m_3} \gd(m_1,m_2,m_3) \webb(x_{m_3}) v_{m_3} \\
&- \sum_{m_3} \gd(m_2,m_3) \wcov(x_{m_1},x_{m_2})v_{m_3}\\
& - \sum_{m_3} \gd(m_1,m_3) \wcov(x_{m_1},x_{m_2}) v_{m_3}\\
& - \sum_{m_3} \gd(m_1,m_2) \wcov(x_{m_1},x_{m_3})v_{m_3} \\
&= \sum_{m_3} \wcum(x_{m_1},x_{m_2},x_{m_3}) v_{m_3} + 2 \gd(m_1,m_2) \webb(x_{m_1}) v_{m_1} \\
&-  \wcov(x_{m_1},x_{m_2}) v_{m_2} -  \wcov(x_{m_1},x_{m_2}) v_{m_1} 
- \gd(m_1,m_2) \sum_{m_3} \wcov(x_{m_1},x_{m_3}) v_{m_3}.
\end{ma}
This gives the following for the expression $\wh{W}\wh{T}(v)\wh{W}^{\top}$:
\begin{ma}
\sbra{\wh{W}\wh{T}(v)\wh{W}^{\top}}_{k_1k_2} &= \wh{W}{k_1}^{\top} \wh{T}(v) \wh{W}_{k_2} \\
&= \sum_{m_1,m_2,m_3} \wh{\cum}(x_{m_1},x_{m_2},x_{m_3}) v_{m_3} \wh{W}_{k_1m_1} \wh{W}_{k_2m_2}\\
&+ 2\sum_{m_1,m_2} \gd(m_1,m_2) \wh{\ebb} (x_{m_1}) v_{m_1} \wh{W}_{k_1m_1} \wh{W}_{k_2m_2} \\
&- \sum_{m_1,m_2} \wh{\cov}(x_{m_1},x_{m_2}) v_{m_2} \wh{W}_{k_1m_1} \wh{W}_{k_2m_2} \\
& - \sum_{m_1,m_2} \wh{\cov}(x_{m_1},x_{m_2}) v_{m_1} \wh{W}_{k_1m_1} \wh{W}_{k_2m_2}\\
& - \sum_{m_1,m_3} \wh{\cov}(x_{m_1},x_{m_3}) v_{m_3} \wh{W}_{k_1m_1} \wh{W}_{k_2m_1},
\end{ma}
where $\wh{W}_k$ denotes the $k$-th row of $\wh{W}$ as a column vector.
By further plugging in the expressions~\eqref{cum:empirical} for the unbiased finite sample estimates of $\wcov$ and $\wcum$, we further get
\begin{ma}
\sbra{\wh{W}\wh{T}(v)\wh{W}^{\top}}_{k_1k_2}&= \frac{N}{(N-1)(N-2)} \sum_n \inner{\wh{W}_{k_1},x_n-\webb(x)} \inner{\wh{W}_{k_2},x_n-\webb(x)} \inner{v, x_n - \webb(x)} \\
&+ 2\sum_m \webb(x_m) v_m \wh{W}_{k_1m} \wh{W}_{k_2m} \\
&- \frac{1}{N-1} \sum_n \inner{\wh{W}_{k_1}, x_n - \webb(x)} \inner{v\circ \wh{W}_{k_2}, x_n - \webb(x)} \\
&- \frac{1}{N-1} \sum_n \inner{v\circ \wh{W}_{k_1}, x_n - \webb(x)} \inner{\wh{W}_{k_2}, x_n - \webb(x)} \\
&- \frac{1}{N-1} \sum_n \inner{\wh{W}_{k_1} \circ \wh{W}_{k_2}, x_n -\webb(x)} \inner{v, x_n - \webb(x)},
\end{ma} 
where $\circ$ denotes the elementwise Hadamard product.
Introducing
the counts matrix $X\in\rr{M\times N}$ where each element $X_{mn}$ is the count of the $m$-th word in the $n$-th document (note, the matrix $X$ contain the vector $x_n$ in the $n$-th column), we further simplify the above expression
\begin{equation}\label{est:T:dica:old}
\begin{aligned}
\wh{W}\wh{T}(v)\wh{W}^{\top} & = \frac{N}{(N-1)(N-2)} (\wh{W}X)\diag[X^{\top} v](\wh{W}X)^{\top} \\
&+\frac{N}{(N-1)(N-2)}\inner{v,\webb(x)}\sbra{ 2N(\wh{W}\webb(x))(\wh{W}\webb(x))^{\top} - (\wh{W}X)(\wh{W}X)^{\top} } \\
& - \frac{N}{(N-1)(N-2)} \sbra{\wh{W}X(X^{\top} v) (\wh{W}\webb(x))^{\top} + \wh{W}\webb(x)(\wh{W}X(X^{\top} v))^{\top}} \\
&+2\wh{W} \diag[v\circ\webb(x)] \wh{W}^{\top} \\
& - \frac{1}{N-1} \sbra{ (\wh{W}X)(\wh{W}\diag(v)X)^{\top} + (\wh{W}\diag(v) X)(\wh{W}X)^{\top} + \wh{W} \diag[X(X^{\top} v)] \wh{W}^{\top} } \\
& + \frac{N}{N-1} \sbra{ (\wh{W}\webb(x))(\wh{W}\diag[v] \webb(x))^{\top} + (\wh{W}\diag[v] \webb(x))(\wh{W}\webb(x))^{\top}  } \\
& + \frac{N}{N-1} \inner{v,\webb(x)}\wh{W} \diag[\webb(x)] \wh{W}^{\top}.
\end{aligned}
\end{equation}

A more compact way to write down expression~\eqref{est:T:dica:old} is as follows
\begin{equation}
\label{est:T:dica}
\begin{aligned}
\wh{W}\wh{T}(v)\wh{W}^{\top} & = \frac{N}{(N-1)(N-2)} \sbra{ T_1 + \innerp{v,\webb(x)}\rbra{ T_2 - T_3 } - ( T_4 + T_4^{\top} ) } \\
& + \frac{1}{N-1} \sbra{ T_5 + T_5^{\top} - T_6 - T_6^{\top} +
\wh{W} \diag(a) \wh{W}^{\top} 
 }, 
\end{aligned}
\end{equation}
where
\begin{ma}
T_1 &= (\wh{W}X)\diag[X^{\top} v](\wh{W}X)^{\top}, \\
T_2 &= 2N(\wh{W}\webb(x))(\wh{W}\webb(x))^{\top}, \\
T_3 &= (\wh{W}X)(\wh{W}X)^{\top}, \\
T_4 &= \wh{W}X(X^{\top} v) (\wh{W}\webb(x))^{\top}, \\
T_5 &= (\wh{W}X)(\wh{W}\diag(v)X)^{\top}, \\
T_6 &= (\wh{W}\diag(v) \webb(x))(\wh{W}\webb(x))^{\top}, \\
a &= 2(N-1)[v\circ \webb(x)] + \innerp{v,\webb(x)}\webb(x) - X(X^{\top} v).
\end{ma}

\subsubsection{Computational complexity of the GP/DICA T-cumulant estimator~\eqref{est:T:dica}}\label{sec:complexity-t-cumulant}
When computing the T-cumulant $P$ times with the formula above, the following terms are dominant: $O(RNK) + O(NK^2) + O(MK)$, where $R$ is the largest number of unique words (non-zero counts) in a document over the corpus. In practice, almost always $K < M < N$, which gives the overall complexity of $P$ computations of the estimator~\eqref{est:T:dica} to be equal to $O(PRNK) + O(PNK^2)$.

\subsubsection{Computation of the finite sample estimators of the LDA moments} \label{sec:lda:empirical}
In this section, we present efficient formulas for computation of the finite sample estimate (see Appendix~\ref{sec:lda:fs} for the definition of $\wh{T}$) of $\wh{W}\wh{T}(v)\wh{W}^{\top}$ for the LDA model. Note that the construction of the sample estimator $\wh{W}$ of a whitening matrix $W$ is discussed in Appendix~\ref{sec:whitening}). The computation of $\wh{S}$~\eqref{S:lda:fs} is straightforward.
This approach to efficient implementation was discussed in~\cite{AnaEtAl2014}, however, to the best of our knowledge, the final expressions were not explicitly stated before. All derivations are straightforward, but quite tedious. 

By analogy with the GP/DICA case, a projection~\eqref{projT} of the tensor $\wh{T}\in\rr{M\times M\times M}$~\eqref{T:lda:fs} onto some vector $v\in\rr{M}$ in the LDA is
\begin{ma}
&\sbra{\wh{T}(v)}_{m_1m_2} = \sum_{m_3=1}^M\sbra{\wh{M}_3}_{m_1m_2m_3} v_{m_3} +\frac{2c_0^2}{(c_0+1)(c_0+2)}\sum_{m_3}[\wh{M}_1]_{m_1}[\wh{M}_1]_{m_2}[\wh{M}_1]_{m_3}v_{m_3} \\
&- \frac{c_0}{c_0+2} \sum_{m_3=1}^M
\sbra{\webb(w_{\ell_1}\tp w_{\ell_2} \tp \wh{M}_1)
+\webb(w_{\ell_1}\tp \wh{M}_1 \tp w_{\ell_3}) 
+ \webb(\wh{M}_1\tp w_{\ell_2}\tp w_{\ell_3})}_{m_1m_2m_3} v_{m_3}.
\end{ma}
Plugging in the expression~\eqref{m3final} for an unbiased sample estimate $\wh{M_3}$ of $M_3$, we get
\begin{ma}
\sbra{\wh{T}(v)}_{m_1m_2} 
&= \frac{1}{N} \sumn [\delta_3]_n \rbra{x_{nm_1}x_{nm_2}\inner{x_n,v} + 2\sum_{m_3}\gd(m_1,m_2,m_3)x_{nm_3}v_{m_3}} \\
& - \frac{1}{N} \sumn [\delta_3]_n \sum_{m_3=1}^M \sbra{\sum_{i,j=1}^M x_{ni}x_{nj}\rbra{ e_i\tp e_i \tp e_j +e_i\tp e_j \tp e_i + e_i \tp e_j \tp e_j}}_{m_1m_2m_3} v_{m_3} \\
& + \frac{2c_0^2}{(c_0+1)(c_0+2)} [\wh{M}_1]_{m_1}[\wh{M}_1]_{m_2} \inner{\wh{M}_1,v}\\
& - \frac{c_0}{c_0+2} \rbra{[\wh{M}_2]_{m_1m_2} \inner{\wh{M}_1,v} + \sum_{m_3=1}^M \rbra{ [\wh{M}_2]_{m_1m_3}[\wh{M}_1]_{m_2}v_{m_3} + [\wh{M}_2 ]_{m_2m_3} [\wh{M}_1]_{m_1}v_{m_3} } },
\end{ma}
where $e_1$, $e_2$, \dots, $e_M$ denote the canonical vectors of $\rr{M}$ (i.e., the columns of the identity matrix $I_M$). 
Further, this gives the following for the expression $\wh{W}\wh{T}(v)\wh{W}^{\top}$:
\begin{align}
\nonumber \sbra{\wh{W}\wh{T}(v)\wh{W}^{\top}}_{k_1k_2} &= \frac{1}{N} \sumn [\delta_3]_n \rbra{\inner{x_n,v}\inner{x_n,\wh{W}_{k_1}}\inner{x_n,\wh{W}_{k_2}} +2\summ x_{nm}v_m \wh{W}_{k_1m}\wh{W}_{k_2m}} \\
\nonumber &- \frac{1}{N} \sumn \delta_{3n} \sum_{i,j=1}^M x_{ni}x_{nj}\rbra{ \wh{W}_{k_1i}\wh{W}_{k_2i}v_j + \wh{W}_{k_1i}\wh{W}_{k_2j}v_i + \wh{W}_{k_1i}\wh{W}_{k_2j}v_j} \\
\nonumber &- \frac{c_0}{c_0+2} \rbra{ \inner{\wh{W}_{k_1},\sbra{\wh{M}_2}\wh{W}_{k_2}} + \inner{\wh{W}_{k_1},\wh{M}_2v}\inner{\wh{M}_1\wh{W}_{k_2} } +  \inner{\wh{W}_{k_2},\wh{M}_2 v}\inner{\wh{M}_1,\wh{W}_{k_1}} } \\
\nonumber &+ \frac{2c_0^2}{(c_0+1)(c_0+2)}\inner{\wh{M}_1,\wh{W}_{k_1}} \inner{\wh{M}_1,\wh{W}_{k_2}}\inner{\wh{M}_1,v},
\end{align}
where $\wh{W}_{k}$ denotes the $k$-th row of $\wh{W}$ as a column-vector.
This  further simplifies to
\begin{equation}
\begin{aligned}
 \wh{W}\wh{T}(v)\wh{W}^{\top} &= \frac{1}{N} (\wh{W}X)\diag\sbra{ (X^{\top} v)\circ \delta_3} (\wh{W}X)^{\top} \\
 & + \frac{1}{N} \wh{W}\diag\sbra{2[(X\delta_3)\circ v]-X[(X^{\top} v)\circ \delta_3] } \wh{W}^{\top}  \\
 &-\frac{1}{N} (\wh{W}\diag[v]X)\diag[\delta_3](\wh{W}X)^{\top} \\
 &-\frac{1}{N}  (\wh{W}X)\diag[\delta_3] (\wh{W}\diag[v]X)^{\top} \\
 &-\frac{c_0}{c_0+2}\sbra{\inner{\wh{M}_1,v}(\wh{W}\wh{M}_2\wh{W}^{\top}) + (\wh{W}(\wh{M}_2v))(\wh{W}\wh{M}_1)^{\top} + (\wh{W}\wh{M}_1)(\wh{W}(\wh{M}_2v))^{\top}}  \\
 & + \frac{2c_0^2}{(c_0+1)(c_0+2)}\inner{\wh{M}_1,v}(\wh{W}\wh{M}_1) (\wh{W}\wh{M}_1)^{\top}.
\end{aligned}
\end{equation}
A more compact representation gives:
\begin{equation}
\label{est:T:lda}
\begin{aligned}
 \wh{W}\wh{T}(v)\wh{W}^{\top} &= \frac{1}{N} \sbra{ T_1 + T_2 -T_3 - T_3^{\top}}  
 -\frac{c_0}{c_0+2}\sbra{\innerp{\wh{M}_1,v}(\wh{W}\wh{M}_2\wh{W}^{\top}) + T_4 + T_4^{\top}}  \\
&  + \frac{2c_0^2}{(c_0+1)(c_0+2)}\innerp{\wh{M}_1,v}(\wh{W}\wh{M}_1) (\wh{W}\wh{M}_1)^{\top},
\end{aligned}
\end{equation}
where
\begin{ma}
T_1 & = (\wh{W}X)\diag\sbra{ (X^{\top} v)\circ \delta_3} (\wh{W}X)^{\top}, \\
T_2 & = \wh{W}\diag\sbra{2[(X\delta_3)\circ v]-X[(X^{\top} v)\circ \delta_3] } \wh{W}^{\top}, \\
T_3 & = [\wh{W}\diag(v)X]\diag(\delta_3)(\wh{W}X)^{\top}, \\
T_4 & = [\wh{W}(\wh{M}_2v)](\wh{W}\wh{M}_1)^{\top}.
\end{ma}

\subsubsection{Computational complexity of the LDA T-moment estimator~\eqref{est:T:lda}}\label{sec:complexity-t-moment}
By analogy with Appendix~\ref{sec:complexity-t-cumulant}, the computational complexity of the T-moment is $O(RNK) + O(NK^2)$. However, in practice we noticed that the computation of~\eqref{est:T:dica} is slightly faster for larger datasets than the computation of~\eqref{est:T:lda} (although the code for both was equally well optimized). This means that the constants in $O(RNK)+O(NK^2)$ for the LDA T-moment are, probably, slightly larger than for the GP/DICA T-cumulant.

\subsubsection{Estimation of the model parameters for GP/DICA model} \label{estimation-model-parameters}
Below we briefly discuss the recovery of the model parameters for the GP/DICA and LDA models from 
a joint diagonalization matrix $A \in \rr{K \times M}$ estimated in Algorithm~\ref{alg:jd}. This matrix has the property that $A D$ should be approximately diagonal up to a permutation of the columns of $D$. The standard approach~\cite{AnaEtAl2012} of taking the pseudo-inverse of $A$ to get an estimate of the topic matrix $D$ has a problem that it does not preserve the simplex constraint of the topics (in particular, the non-negativity of $\wt{D}$). Due to the space constraints, we do not discuss this issue here, but we observed experimentally that this can potentially significantly deteriorate performance of all moment matching algorithms for topic models considered in this paper. We made an attempt to solve this problem by integrating the non-negativity constraint into the Jacobi-updates procedure of the orthogonal joint diagonalization algorithm, but the obtained results did not lead to any significant improvement. Therefore, in our experiments for both GP/DICA cumulants and LDA moments, we estimate the topic matrix by thresholding the negative values of the pseudo-inverse of $A$: 
$$
\wh{d}_k := \tau_k\max(0,[A\pinv]_{:k})/\normp{\max(0,[A\pinv]_{:k})}_1,
$$ 
where $[A\pinv]_{:k}$ is the $k$-th column of the pseudo-inverse $A\pinv$ of $A$, and $\tau_k=\rpm 1$ set to $-1$ if $[A\pinv]_{:k}$ has more negative than positive values. This might not be the best option, and we leave this issue for the future research.

To estimate the parameters for the prior distribution over the topic intensities $\ga_k$ for the DICA model~\eqref{dica}, we use the diagonalized form of the projected tensor from~\eqref{orthT} and relate it to the output diagonal elements $a_p$ for the $p$-th projection:
\begin{equation} \label{eq:ap_tk}
[a_p]_k = \wt{t}_k \innerp{z_k,u_p} = \frac{t_k}{s_k^{3/2}}  \innerp{z_k,u_p} =
\frac{\cum(\ga_k,\ga_k,\ga_k)}{[\var(\ga_k)]^{3/2}} \inner{\tau_k\wt{d}_k,W^{\top}u_p},
\end{equation}
where $\wt{d}_k = \tau_k \max(0,[A\pinv]_{:k})$.
This formula is valid for any prior on $\ga_k$ in the DICA model. For the GP model~\eqref{gp} where $\ga_k \sim \gam(c_k,b)$, we have that $\var(\ga_k) = \frac{c_k}{b^2}$ and  $\cum(\ga_k,\ga_k,\ga_k) = \frac{2 c_k}{b^3}$, and thus $ \wt{t}_k  = \frac{2}{\sqrt{c_k}}$, which enables us to estimate $c_k$. Plugging this value of $\wt{t}_k$ in~\eqref{eq:ap_tk}, and solving for $c_k$ gives the following expression:
$$
c_k = \frac{4 \inner{\wt{d}_k,W^{\top}u_p}^2}{[a_p]_k^2}.
$$
By replacing the quantities on the RHS with their estimated ones, we get one estimate for $c_k$ per projection. We use as our final estimate the average estimate over the projections:
\begin{equation} \label{eq:ck_estimate}
\wh{c}_k := \frac{1}{P} \sum_{p=1}^P  \frac{4 \inner{\wt{d}_k,\wh{W}^{\top}u_p}^2}{[a_p]_k^2}.
\end{equation}
Reusing the properties of the length of documents for the GP model as described in Appendix~\ref{sec:L:gp}, we finally use the following estimates for rate parameter $b$ of the gamma distribution:
\begin{equation} \label{eq:b_estimate}
 \wh{b} := \frac{\wh{c}_0} {\wh{L}},
\end{equation}
where $\wh{c}_0 := \sum_k \wh{c}_k$ and $\wh{L}$ is the average document length in the corpus.

By analogy, similar formulas for the estimation of the Dirichlet parameter $c$ of the LDA model can be derived and are a straightforward extension of the expression in~\cite{AnaEtAl2012}.

\subsection{Appendix. Complexity of algorithms and details on the experiments}

\subsubsection{Code and complexity}\label{sec:code-and-compleixty}

Our (mostly Matlab) implementations of the diagonalization algorithms (JD, Spec, and TPM) for both the GP/DICA cumulants and LDA moments are available online.\footnote{\url{https://github.com/anastasia-podosinnikova/dica-light}} Moreover, all datasets and the code for reproducing our experiments are available.\footnote{\url{https://github.com/anastasia-podosinnikova/dica}}
To our knowledge, no efficient implementation of these algorithms was available for LDA. Each experiment was run in a single thread.

The bottleneck for the spectral, JD, and TPM algorithms is the computation of the cumulants/moments.
However, the expressions~\eqref{est:T:dica} and~\eqref{est:T:lda} provide efficient formulas for fast computation of the GP/DICA cumulants and LDA moments ($O(RNK + NK^2)$, where $R$ is the largest number of non-zeros in the count vector $x$ over all documents, see Appendix~\ref{sec:complexity-t-cumulant} and~\ref{sec:complexity-t-moment}), which makes even the Matlab implementation fast for large datasets. Since all diagonalization algorithms (spectral, JD, TPM) perform the whitening step once,
it is sufficient to compare their complexities by the number of times the cumulants/moments are computed.

\textbf{Spectral.} The spectral algorithm estimates the cumulants/moments only once leading to $O(NK(R+K))$ complexity and, therefore, is the fastest.

\textbf{JD.}
For JD, rather than estimating $P$ cumulants/moments separately, one can jointly estimate these values by precomputing and reusing some terms (e.g., $W X$).
However, the complexity is still $O(PNK(R + K))$, although in practice it is sufficient to have $P=K$ or even smaller.

\textbf{TPM.} For TPM some parts of the cumulants/moments can also be precomputed, but as TPM normally does many more iterations than $P$, it can be significantly slower. 
In general, the complexity of TPM can be significantly influenced by the initialization of the parameters of the algorithm. There are two main parameters: $L_{tpm}$ is the number of random restarts within one deflation step and $N_{tpm}$ is the maximum number of iterations for each of $L_{tpm}$ random restarts (different from $N$ and $L$). Some restarts converge very fast (in much less than $N_{tpm}$ iterations), while others are slow. Moreover, as follows from theoretical results~\cite{AnaEtAl2014} and, as we observed in practice, the restarts which converge to a good solution converge fast, while slow restarts, normally, converge to a worse solution. Nevertheless, in the worst case, the complexity is $O(N_{tpm} L_{tpm}NK(R+K))$.

Note that for the experiment in Figure~\ref{plots:diag}, $L_{tpm}=10$ and $N_{tpm} = 100$ and the run with the best objective is chosen. We believe that these values are reasonable in a sense that they provide a good accuracy solution ($\varepsilon = 10^{-5}$ for the norm of the difference of the vectors from the previous and the current iteration) in a little number of iterations, however, they may not be the best ones. 

\textbf{JD implementation.} 
For the orthogonal joint diagonalization algorithm, we implemented a faster C++ version of the previous Matlab implementation\footnote{\url{http://perso.telecom-paristech.fr/~cardoso/Algo/Joint_Diag/joint_diag_r.m}} by J.-F.~Cardoso. Moreover, the orthogonal joint diagonalization routine can be initialized in different ways: (a) with the $K\times K$ identity matrix or (b) with a random orthogonal $K\times K$ matrix. We tried different options and in nearly all cases the algorithm converged to the same solution, implying that initialization with the identity matrix is sufficient.

\textbf{Whitening matrix.} For the large vocabulary size $M$,  computation of a whitening matrix can be expensive (in terms of both memory and time).
One possible solution would be to reduce the vocabulary size with, e.g., TF-IDF score, which is a standard practice in the topic modeling context.
Another option is using a stochastic eigendecomposition (see, e.g.,~\citesup{HalEtAl2011}) to approximate the whitening matrix.

\textbf{Variational inference.} For variational inference, we used the code of D.~Blei and modified it for the estimation of a non-symmetric Dirichlet prior $c$, which is known to be important~\citesup{WalEtAl2009b}. The default values of the tolerance/maximum number of iterations parameters are used for variational inference. The computational complexity of one iteration for one document of the variational inference algorithm is $O(RK)$, where $R$ is the number of non-zeros in the count vector for this document, which is then performed a significant number of times for each document. 

\subsubsection{Runtimes of the algorithms}\label{sec:runtimes}
In Table~\ref{tab:runtimes}, we present the running times of the algorithms from Section~\ref{sec:diagcmp}. 
\begin{table}[t!]
\centering
\begin{tabular}{| l | l l l |} \hline
           & min   & mean  & max \\ \hline
 JD-GP     & 148   & 192   & 247 \\
 JD-LDA    & 252   & 284   & 366 \\ \hline
 JD(k)-GP  & 157   & 190   & 247 \\
 JD(k)-LDA & 264   & 290   & 318 \\ \hline
 JD(f)-GP  & 1628  & 1846  & 2058 \\
 JD(f)-LDA & 2545  & 2649  & 2806 \\ \hline
 Spec-GP   & 101   & 107   & 111 \\
 Spec-LDA  & 107   & 140   & 193 \\ \hline
 TPM-GP    & 1734  & 2393  & 2726 \\
 TPM-LDA   & 12723 & 16460  & 19356 \\ \hline
\end{tabular}
\caption{The running times in seconds of the algorithms from Figure~\ref{plots:diag}, corresponds to the case when $N=50,000$. Each algorithm was run $5$ times, so the times in the table display the minimum (min), mean, and maximum (max) time. }
\label{tab:runtimes}
\end{table} 
JD and JD(k) are significantly faster than JD(f) as expected, although the performance in terms of the $\ell_1$-error is nearly the same for all of them. This indicates that preference should be given to the JD or JD(k) algorithms.

The running time of all LDA-algorithms is higher than the one of the GP/DICA-algorithms. This indicates that the computational complexity of the LDA-moments is slightly higher than the one of the GP/DICA-cumulants (compare, e.g., the times for the spectral algorithm which almost completely consist of the computation of the moments/cumulants). Moreover, the runtime of TPM-LDA is significantly higher (half an hour vs. several hours) than the one of TPM-GP/DICA. This can be explained by the fact that the LDA-moments have more noise than the GP/DICA-cumulants and, hence, the algorithm is slower. Interestingly, all versions of JD algorithm are not that sensitive to noise. 

Computation of a whitening matrix is roughly 30 sec (this time is the same for all algorithms and is included in the numbers above). 

\subsubsection{Initialization of the parameter c\textsubscript{0} for the LDA moments}\label{sec:initialization-c0}
The construction of the LDA moments requires the parameter $c_0$, which is not trivial to set in the unsupervised setting of topic modeling, especially taking into account the complexity of the evaluation for topic models~\cite{WalEtAl2009}. For the semi-synthetic experiments, the true value of $c_0$ is provided to the algorithms. It means that the LDA moments, in this case, have access to some oracle information, which in practice is never available. For real data experiments, $c_0$ is set to the value obtained with variational inference. The experiments in Appendix~\ref{sec:app:c0lda} show that this choice was somewhat important. However, this requires more thorough investigation.

\subsubsection{The LDA moments vs parameter c\textsubscript{0}}\label{sec:app:c0lda}
\begin{figure}[t]
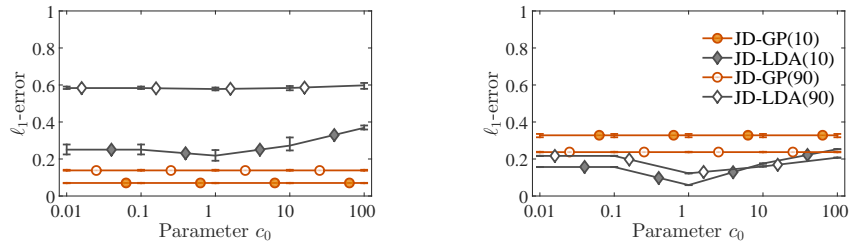

\centering
\begin{tabular}{cccc}
\includegraphics[width=.36\columnwidth]{fig_5_left_l1.eps} 
 & 
 
 &
 
 &
\includegraphics[width=.36\columnwidth]{fig_5_right_l1.eps} 
\end{tabular}
\caption{ Performance of the LDA moments depending on the parameter $c_0$. $D$ and $c$ are learned from the AP dataset for $K=10$ and $K=50$ and true $c_0=1$. JD-GP(10) for $K=10$ and JD-GP(50) for $K=50$. Number of sampled documents $N=20,000$. For the error bars, each dataset is resampled 5 times. Data \textbf{(left)}: \textit{GP} sampling; \textbf{(right)}: \textit{LDAfix($200$)} sampling.  
\textit{Note}: a smaller value of the $\ell_1$-error is better.  }
\label{fig:4}
\end{figure}

In this section, we experimentally investigate dependence of the LDA moments on the parameter~$c_0$.
 In Figure~\ref{fig:4}, the joint diagonalization algorithm with the LDA moment is compared for different values of $c_0$ provided to the algorithm. The data is generated similarly to Figure~\ref{plot:moms}. The experiment indicates that the LDA moments are somewhat sensitive to the choice of $c_0$. For example, the recovery $\ell_1$-error doubles when moving from the correct choice $c_0=1$ to the plausible alternative $c_0 = 0.1$ for $K=10$ on the \textit{LDAfix($200$)} dataset (JD-LDA(10) line on the right of Figure~\ref{fig:4}). 

\subsubsection{Comparison of the \texorpdfstring{$\mathbf{\ell_1}$}{L1}- and \texorpdfstring{$\mathbf{\ell_2}$}{L2}-errors}
\begin{figure}[t]
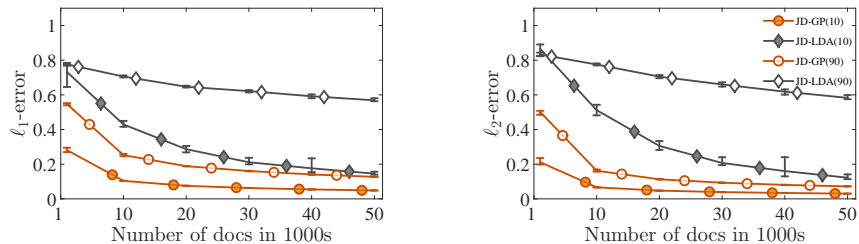

\centering
\begin{tabular}{cccc}
\includegraphics[width=.36\columnwidth]{fig_6_left.eps} 
 & 
 
 &
 
 &
\includegraphics[width=.36\columnwidth]{fig_6_right.eps} 
\end{tabular}
\caption{Comparison of the $\ell_1$- and $\ell_2$- errors on the NIPS semi-synthetic dataset as in Figure~\ref{plot:moms} (top, left). The $\ell_2$ norms of the topics were normalized to [0,1] for the computation of the $\ell_2$ error. }
\label{l1vsl2}
\end{figure}

The sample complexity results~\cite{AnaEtAl2012} for the spectral algorithm for the LDA moments allow straightforward extension to the GP/DICA cumulants, if the results from Proposition~\ref{sample-complexity} are taken into account. The analysis is, however, in terms of the $\ell_2$-norm. Therefore, in Figure~\ref{l1vsl2}, we provide experimental comparison of the $\ell_1$- and $\ell_2$-errors to verify that they are indeed behaving similarly.

\subsubsection{Evaluation of the real data experiments}\label{sec:evaluation-real}
For the evaluation of topic recovery in the real data case, we use an approximation of the log-likelihood for held out documents as the metric. The approximation is computed using a Chib-style method as described by~\cite{WalEtAl2009} using the implementation by the authors.\footnote{\url{http://homepages.inf.ed.ac.uk/imurray2/pub/09etm}}
Importantly, this evaluation methods is applicable for both the LDA model as well as the GP model. Indeed, as it follows from Section~\ref{sec:2} and Appendix~\ref{sec:ldaproof2}, the GP model is equivalent to the LDA model when conditioning on the length of a document $L$ (with the same $c_k$ hyper parameters), while the LDA model does not make any assumption on the document length. For the test log-likelihood comparison, we thus treat the GP model as a LDA model (we do not include the likelihood of the document length).

\subsubsection{More on the real data experiments}\label{sec-expsup-real}
\begin{figure}[!h]
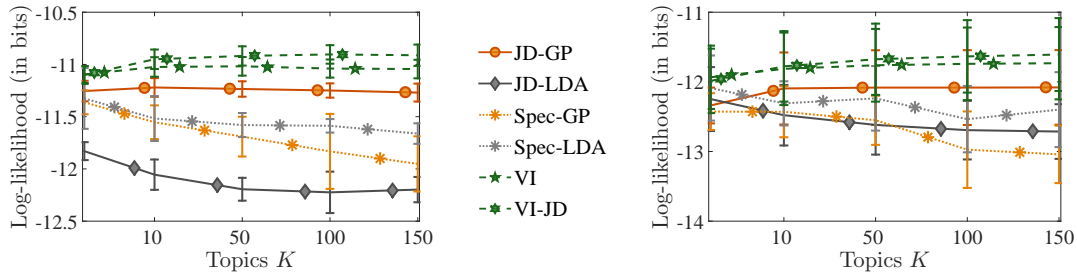

\begin{center}
\begin{tabular}{ccc}
\includegraphics[width=.41\columnwidth]{fig_3_right_kos.eps} 
 & 
\includegraphics[width=.14\columnwidth]{real_legend.eps} 
 & 
\includegraphics[width=.41\columnwidth]{real_nips.eps} 
\end{tabular}
\vspace{-1em}
\caption{ 
Experiments with real data. \emp{Left:} the KOS dataset. \emp{Right:} the NIPS dataset. \textit{Note}: a higher value of the log-likelihood is better.
}
\label{lastfigure-2}
\end{center}
\end{figure}

The detailed experimental setup is as follows.
Each dataset is separated into 5 training/evaluation pairs, where the documents for evaluation are chosen randomly and non-repetitively  among the folds (600 documents are held out for KOS; 400 documents are held out for AP; 450 documents are held out for NIPS). Then, the model parameters are learned  for a different number of topics. The evaluation of the held-out documents is performed with averaging over 5 folds. In Figure~\ref{lastfigure} and Figure~\ref{lastfigure-2}, on the y-axis, the predictive log-likelihood in bits averaged per token is presented. 

In addition to the experiments with AP and KOS in Figure~\ref{lastfigure}, we demonstrate one more experiment with the NIPS dataset in Figure~\ref{lastfigure-2} (right).

Note that, as the LDA moments require at least 3 tokens in each document, $1$ document from the NIPS dataset and 3 documents from the AP dataset, which did not fulfill this requirement, were removed.

Importantly, we observed that VI when initialized with the output of the JD-GP is consistently better in terms of the predictive log-likelihood.
Therefore, the new algorithm can be used for more clever initialization of other LDA/GP inference methods.

We also observe that the joint diagonalization algorithm for the LDA moments is worse than the spectral algorithm. This indicates that the diagonal structure~\eqref{diagS:lda} and~\eqref{diagT:lda} might not be present in the sample estimates~\eqref{S:lda:fs} and~\eqref{T:lda:fs} due to either model misspecification or to finite sample complexity issues.

\bibliographysup{litsup}
\bibliographystylesup{plain}

\end{document}